%% file: shape_adaptor_arxiv.tex
\documentclass[11pt]{article}
\usepackage{graphicx, fleqn, tabularx, wrapfig}
\usepackage{amsthm, amsmath, textcomp, amssymb}
\usepackage[letterpaper, margin=1.2in]{geometry}
\usepackage[ruled, linesnumbered]{algorithm2e}
\usepackage[makeroom]{cancel}
\usepackage[table]{xcolor}
\usepackage{enumerate}
\usepackage[T1]{fontenc}
\usepackage{mathpazo}
\usepackage{tgpagella}
\usepackage{parskip}
\usepackage{subcaption}
\usepackage{mathtools}
\usepackage{booktabs}
\usepackage{makecell}
\usepackage{hyperref}
\usepackage{multirow}
\usepackage{tabularx}
\usepackage{authblk}
\usepackage{nicefrac,xfrac}
\usepackage[export]{adjustbox}
\usepackage[colorinlistoftodos]{todonotes}
\usepackage{listings}
\usepackage{xcolor}
\usepackage{pgfplots}
\usepackage[all]{nowidow}

\pgfplotsset{compat=newest}
\usepgfplotslibrary{groupplots}
\usepgfplotslibrary{dateplot}
\usetikzlibrary{arrows.meta}

\definecolor{coolgrey}{RGB}{157,157,157}
\definecolor{lightgrey}{RGB}{235,238,238}
\definecolor{lightteal}{RGB}{198,211,222}

\definecolor{cyan}{RGB}{136, 204, 238}
\definecolor{teal}{RGB}{68, 170, 153}
\definecolor{sand}{RGB}{221, 204, 119}
\definecolor{rose}{RGB}{204, 102, 119}

\definecolor{red}{RGB}{250, 94, 91}
\definecolor{orange}{RGB}{255, 200, 63}
\definecolor{yellow}{RGB}{254, 239, 109}

\newcommand*{\tikzbullet}[2]{%
   \setbox0=\hbox{\strut}%
   \begin{tikzpicture}
     \useasboundingbox (-.2em,0) rectangle (.2em,\ht0);
     \filldraw[draw=#1,fill=#2] (0,0.5\ht0) circle[radius=.2em];
   \end{tikzpicture}%
}

\newcolumntype{L}[1]{>{\raggedright\let\newline\\\arraybackslash\hspace{0pt}}m{#1}}
\newcolumntype{C}[1]{>{\centering\let\newline\\\arraybackslash\hspace{0pt}}m{#1}}
\newcolumntype{R}[1]{>{\raggedleft\let\newline\\\arraybackslash\hspace{0pt}}m{#1}}
\newcolumntype{Y}{>{\centering\arraybackslash}X}

\makeatletter
\newcommand\notsotiny{\@setfontsize\notsotiny{7}{8}}
\makeatother

\lstdefinestyle{mystyle}{
    commentstyle=\color{codegreen},
    keywordstyle=\color{magenta},
    numberstyle=\tiny\color{codegray},
    stringstyle=\color{codepurple},
    basicstyle=\ttfamily\footnotesize,
    breakatwhitespace=false,
    breaklines=true,
    captionpos=b,
    keepspaces=true,
    numbersep=5pt,
    showspaces=false,
    showstringspaces=false,
    showtabs=false,
    tabsize=2,
    frame=single
}

\title{Shape Adaptor: A Learnable Resizing Module}
\author[1]{Shikun Liu\thanks{Corresponding Author: shikun.liu17@imperial.ac.uk.}}
\author[2]{Zhe Lin}
\author[2]{Yilin Wang}
\author[2]{Jianming Zhang}
\author[2]{Federico Perazzi}
\author[1]{\\ Edward Johns}

\affil[1]{Department of Computing, Imperial College London}
\affil[2]{Adobe Research}

\date{}

\begin{document}
\maketitle

\begin{abstract}
  We present a novel resizing module for neural networks: {\it shape adaptor}, a drop-in enhancement built on top of traditional resizing layers, such as pooling, bilinear sampling, and strided convolution. Whilst traditional resizing layers have fixed and deterministic reshaping factors, our module allows for a learnable reshaping factor. Our implementation enables shape adaptors to be trained end-to-end without any additional supervision, through which network architectures can be optimised for each individual task, in a fully automated way. We performed experiments across seven image classification datasets, and results show that by simply using a set of our shape adaptors instead of the original resizing layers, performance increases consistently over human-designed networks, across all datasets. Additionally, we show the effectiveness of shape adaptors on two other applications: network compression and transfer learning. The source code is available at: \url{https://github.com/lorenmt/shape-adaptor}.
\end{abstract}

\section{Introduction}
Deep neural networks have become popular for many machine learning applications, since they provide simple strategies for end-to-end learning of complex representations. However, success can be highly sensitive to network architectures, which places a great demand on manual engineering of architectures and hyper-parameter tuning.

A typical human-designed convolutional neural architecture is composed of two types of computational modules: i) a \textit{normal layer}, such as a stride-1 convolution or an identity mapping, which maintains the spatial dimension of incoming feature maps; ii) a \textit{resizing layer}, such as max/average pooling, bilinear sampling, or stride-2 convolution, which reshapes the incoming feature map into a different spatial dimension. We hereby define the {\it shape} of a neural network as the composition of the feature dimensions in all network layers, and the {\it architecture} as the overall structure formed by stacking multiple normal and resizing layers.

To move beyond the limitations of human-designed network architectures, there has been a growing interest in developing Automated Machine Learning (AutoML) algorithms \cite{hutter2019automated} for automatic architecture design, known as Neural Architecture Search (NAS)  \cite{pmlr-v80-pham18a,cai2018proxylessnas,liu2018darts,liu2018progressive}. However, whilst this has shown promising results in discovering powerful network architectures, these methods still rely heavily on human-designed network shapes, and focus primarily on learning connectivities between layers. Typically, reshaping factors of 0.5 (down-sampling) and 2 (up-sampling) are chosen, and the total number of reshaping layers is defined manually, but we argue that network shape is an important inductive bias which should be directly optimised.

\begin{figure}[t]
  \begin{minipage}[t]{.62\linewidth}
  \centering
  \vspace{0pt}
  \includegraphics[width=\linewidth]{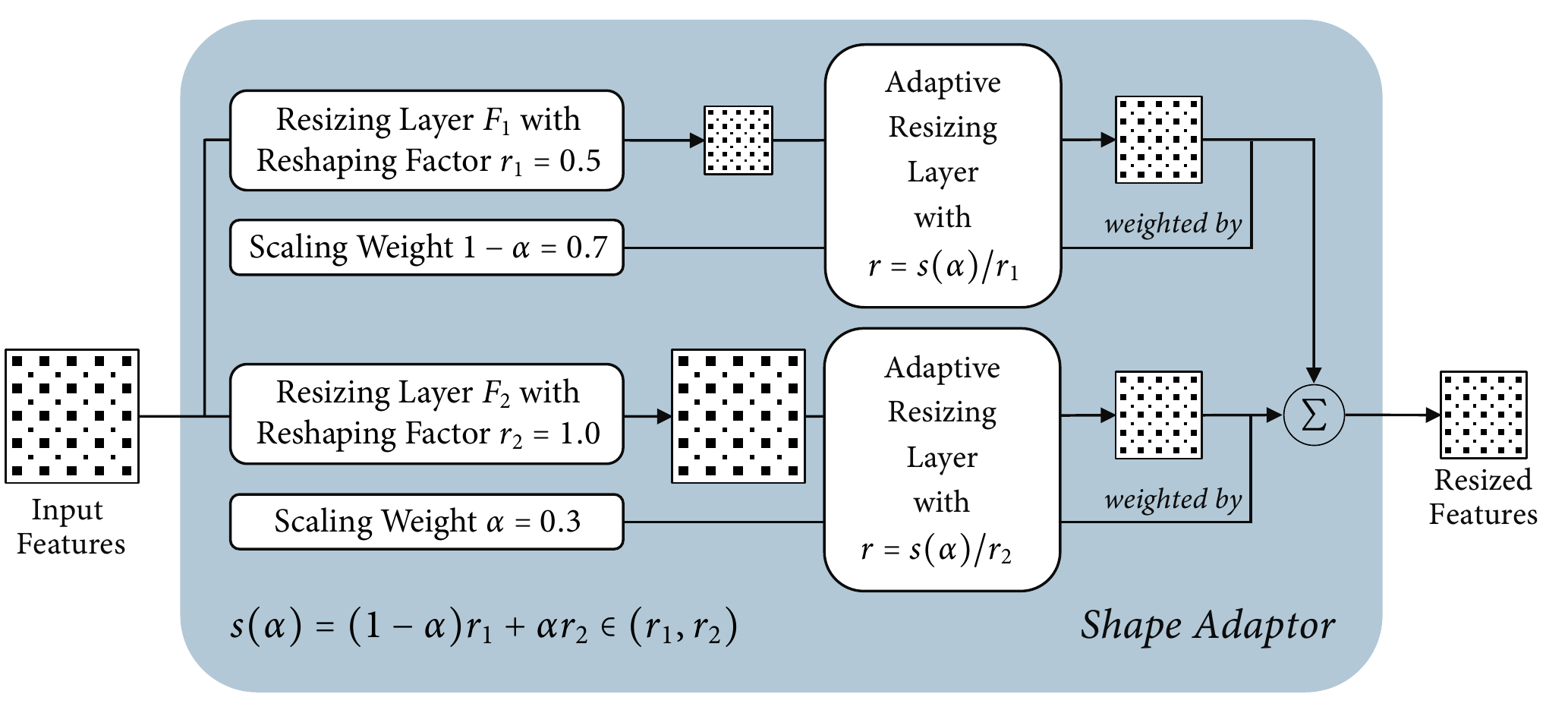}
  \end{minipage}%
  \begin{minipage}[t]{.38\linewidth}
    \scriptsize
    \vspace{0pt}
    \renewcommand{\arraystretch}{1.4}
    \setlength{\tabcolsep}{0.1em}
    \begin{tabularx}{\textwidth}{*{3}{Y}}
      \multicolumn{3}{c}{ResNet-50 on CIFAR-100}\\
      \includegraphics[width=0.9\linewidth, height=1.3\linewidth]{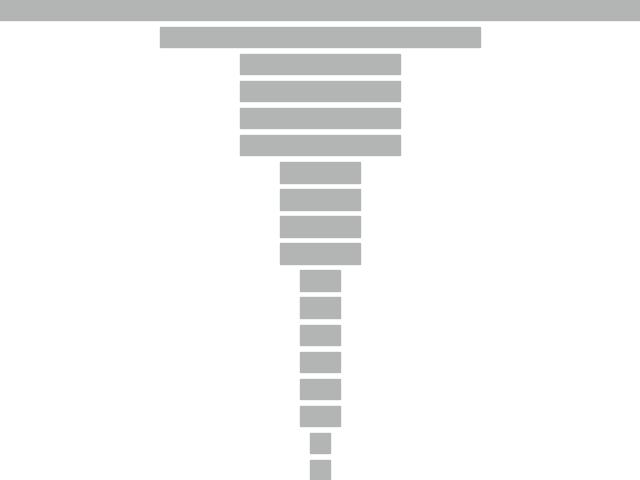} &
      \includegraphics[width=0.9\linewidth, height=1.3\linewidth]{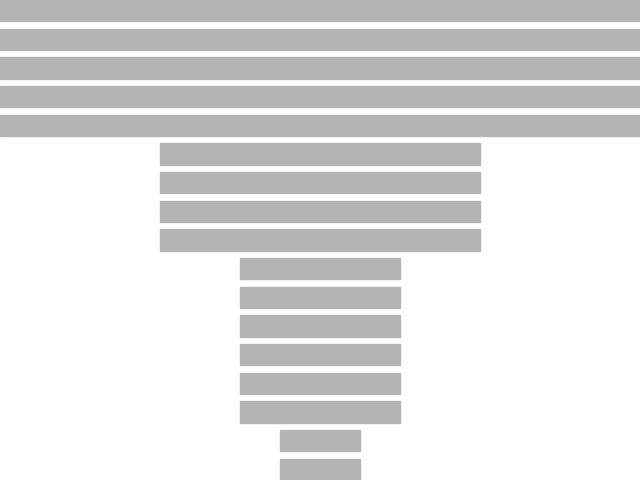} &
      \includegraphics[width=0.9\linewidth, height=1.3\linewidth]{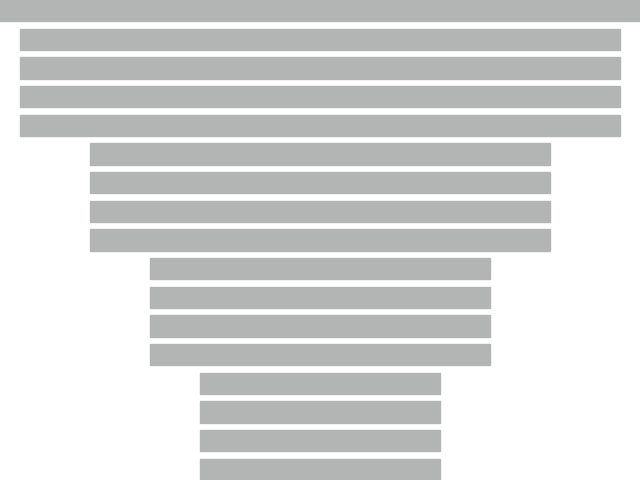} \\
      Acc: 63.08 & Acc: 78.53 & Acc: 80.29 \\
       \makecell{\it  Human\\ \it Designed A}  & \makecell{\it Human\\ \it Designed B} & \makecell{\it Shape\\ \it Adaptors}   \\
    \end{tabularx}
  \end{minipage}
  \caption{Left: Visualisation of a shape adaptor module build on top of two resizing layers. Right: Different network shapes in the exact same network architecture ResNet-50 can result a significantly different performance.}
  \label{fig:intro}
\end{figure}

For example, Figure \ref{fig:intro} Right shows three networks with the exact same design of network structure, but different shapes. For the two human-designed networks \cite{he2016deep}, we see that a ResNet-50 model designed specifically for  CIFAR-100 dataset (Human Designed B) leads to a 15\% performance increase over a ResNet-50 model designed for ImageNet dataset (Human Designed A). The performance can be further improved with the network shape designed by the shape adaptors we will later introduce. Therefore, by learning network shapes rather than manually designing them, a more optimal network architecture can be found. 

To this end, we propose {\it Shape Adaptor}, a novel resizing module which can be dropped into any standard neural network to learn task-specific network shape. A shape adaptor module (see Figure \ref{fig:intro} Left) takes in an input feature, and reshapes it into two intermediate features. Each reshaping operation is done using a standard resizing layer $F_i(x,r_i),i=1,2$, where each resizing layer has a different, pre-defined reshaping factor $r_i$ to reshape feature map $x$. These two reshaping factors then define the search space $(r_1,r_2)$ (assuming $r_1<r_2$) for a shape adaptor module. Finally, the two intermediate features are softly combined with a scalar weighting $1-\alpha$ and $\alpha$ respectively (for $\alpha\in(0,1)$), after reshaping them into the same spatial dimension via a learned reshaping factor in the search space $s(\alpha)\in(r_1,r_2)$. The module's output represents a mixed combination over these two intermediate features, and the scalar $\alpha$ can be learned solely based on the task-specific training loss with stochastic gradient descent, without any additional supervision. Thus, by simply optimising these scaling weights for every shape adaptor, the entire neural architecture is differential and we are able to learn network shape in an automated, end-to-end manner.

We evaluated shape adaptors on seven standard image classification datasets of various complexities. Our results show that shape adaptors can consistently improve human-designed networks, and notably achieve up to 10\% relative performance gains on fine-grained classification datasets. Further experiments show that shape adaptors are robust to initialisations and hyper-parameters, and for a given dataset, they consistently result in the same overall network shape, suggesting that shape adaptors are able to achieve a globally optimal shape. Finally, we further show the effectiveness of shape adaptors in two additional applications: automated neural shape compression, and architecture-level transfer learning.

\section{Related Work \& Background}
\paragraph{Neural Architecture Search}  Neural architecture search (NAS) presents an interesting research direction in AutoML, in automatically discovering an optimal neural structure for a particular dataset, alleviating the hand design of neural architectures which traditionally involves tedious trial-and-error. NAS approaches can be highly computationally demanding, requiring hundreds of thousands of GPU days of search time, due to intensive techniques such as reinforcement learning \cite{zoph2017neural} and evolutionary search \cite{real2019regularized}. Several approaches have been proposed to speed up the search, based on parameter sharing \cite{pmlr-v80-pham18a}, hyper-networks \cite{brock2018smash}, and gradient-based optimisation \cite{liu2018darts}. But despite their promising performance, these approaches come with controversial debate questioning the lack of reproducibility, and sensitivity to initialisations \cite{li2019random,Yu2020Evaluating}. Whilst NAS methods learn network structures based on pre-defined network shapes, shape adaptors are designed in an orthogonal direction, and instead search network shapes in pre-defined network structures. Nevertheless, shape adaptors could be potentially incorporated into modern NAS frameworks, which we consider as future work.

\paragraph{Architecture Pruning \& Compression} Network pruning is another direction towards obtaining optimal network architectures. But instead of searching from scratch as in NAS, network pruning is applied to existing human-design networks and removes redundant neurons and connectivities. Such methods can be based on $\mathcal{L}_0$ regularisation \cite{louizos2018learning}, batch-norm scaling parameters \cite{liu2017learning}, and weight quantization \cite{han2015deep_compression}. As with our shape adaptors, network pruning does not require the extensive search cost of NAS, and can performed alongside regular training. Our shape adaptors can also be formulated as a pruning algorithm, by optimising the network shape within a bounded search space. We provide a detailed explanation of this in Section \ref{sec:autosc}.

\paragraph{Design of Resizing Modules} A resizing module is one of the essential components in deep convolutional network design, and has seen continual modifications to improve performance and efficiency. The most widely used resizing modules are max pooling, average pooling, bilinear sampling, and strided convolutions, which are deterministic, efficient, and simple. But despite their benefits in increasing computational efficiency and providing regularisation, there are two issues with current designs: i) {\it lack of spatial invariance}, and ii) {\it fixed scale}. Prior works focus on improving spatial robustness with a learnable combination between max and average pooling \cite{yu2014mixed,lee2016generalizing}, and with anti-aliased low-pass filters \cite{zhang2019making}. Other works impose regularisation and adjustable inference by stochastically inserting pooling layers \cite{zeiler2013stochastic,kuen2018stochastic}, and sampling different network shapes \cite{zhu2020resizable}. In contrast, shape adaptors solve both problems simultaneously, with a learnable mixture of features in different scales, and with which reshaping factors can be optimised automatically based on the training objective.

\section{Shape Adaptors}
In this section, we introduce the details of the proposed shape adaptor module. We discuss the definition of these modules, and the optimisation strategy used to train them.

\subsection{Formation of Shape Adaptors}
A visual illustration of a shape adaptor module is presented in Figure \ref{fig:intro} Left.  It is a two-branch architecture composed of two different resizing layers $F_i(x,r_i)_{i=1,2}$, assuming $r_1 < r_2$, taking the same feature map $x$ as the input. A resizing layer $F_i$ can be any classical sampling layer, such as max pooling, average pooling, bilinear sampling, or strided convolution, with a fixed reshaping factor $r_i$. Each resizing layer reshapes the input feature map by this factor, which represents the ratio of spatial dimension between the output and input feature maps, and outputs an \textit{intermediate feature}. An adaptive resizing layer $G$ with a learnable reshaping factor is then used to reshape these intermediate features into the same spatial dimension, and combine them with a weighted average to compute the module's output. 

Each module has a learnable parameter $\alpha\in(0,1)$, parameterised by a sigmoid function, which is the only extra learnable parameter introduced by shape adaptors. The role of $\alpha$ is to optimally combine two intermediate features after reshaping them by an adaptive resizing layer $G$. To enable a non-differential reshaping factor in $G$ to be learned, we use a monotone function $s$, which monotonically maps from $\alpha$ into the search space $s(\alpha) \in \mathcal{R}=(r_1,r_2)$, representing the scaling ratio of the module's reshaping operation. With this formulation, a learnble reshaping factor $s(\alpha)$ allows a shape adaptor to reshape at any scale between $r_1$ and $r_2$, rather than being restricted to a discrete set of scales as with typical manually-designed network architectures.

Using this formulation, a shape adaptor module can be expressed as function:
\begin{equation}
{\tt ShapeAdaptor}(x, \alpha, r_{1,2}) =  (1-\alpha) \cdot G\left(F_1(x, r_{1}),\, \frac{s(\alpha)}{r_{1}}\right) + \alpha \cdot G\left(F_2(x,r_{2}),\, \frac{s(\alpha)}{r_{2}}\right),
\end{equation}
with reshaping factor $s(\alpha)$, a {\it monotonic} function which satisfies,
\begin{equation}
    \lim_{\alpha\to 0} s(\alpha)=r_1,\quad \text{and}\quad \lim_{\alpha\to 1} s(\alpha)=r_2.
\end{equation}

We choose our adaptive resizing layer $G$ to be a bilinear interpolation function, which allows feature maps to be resized into any shape. We design module's learnable reshaping factor $s(\alpha)=(r_2-r_1)\alpha +r_1$, a convex combination over these pre-defined reshaping factors, assuming having no prior knowledge on the network shape. 

Each shape adaptor is arranged as a soft and learnable operator to search the optimal reshaping factor $s(\alpha^\ast)=r^\ast\in \mathcal{R}$ over a combination of intermediate reshaped features $F_i(x, r_i)$. Thus, it can also be easily coupled with a continuous approximate of categorical distribution, such as Gumbel SoftMax \cite{jang2017categorical,maddison2017concrete}, to control the softness. This technique is commonly used in gradient-based NAS methods \cite{liu2018darts}, where a categorical distribution is learned over different operations.

The overall shape adaptor module ensures that its reshaping factor $s(\alpha)$ can be updated through the updated scaling weights. Thus, we enable differentiability of $s(\alpha)$ in a shape adaptor module as an approximation from the mapping of the derivative of its learnable scaling weight: $\nabla s(\alpha) \approx s(\nabla \alpha)$. This formulation enables shape adaptors to be easily trained with standard back-propagation and end-to-end optimisation.

In our implementation, we use one resizing layer to maintain the incoming feature dimension (an identity layer), and the other resizing layer to change the dimension. If $F_2$ is the layer which maintains the dimension with $r_2=1$, then a shape adaptor module acts as a learnable down-sampling layer when $0<r_1<1$, and a learnable up-sampling layer when $r_1>1$.

\begin{figure}[t!]
    \centering
    \includegraphics[width=1.0\linewidth]{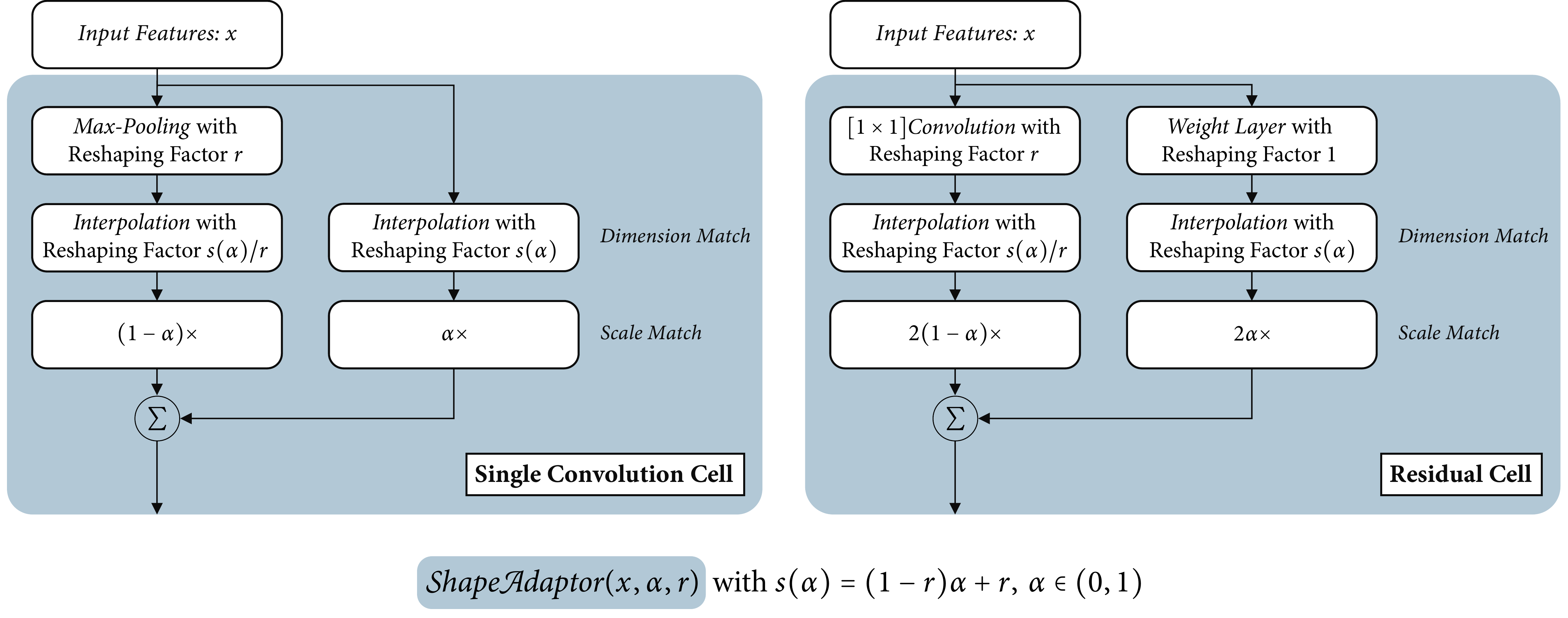}
    \caption{Visualisation of a down-sampling shape adaptor built on a single convolutional cell and a residual cell with a reshaping factor in the range $\mathcal{R}=(r,1)$.}
    \label{fig:structure}
\end{figure}

In Figure \ref{fig:structure}, we illustrate our learnable down-sampling shape adaptor in two commonly used computational modules: a single convolutional cell in VGG-like neural networks \cite{Simonyan15}, and a residual cell in ResNet-like neural networks \cite{he2016deep}. To seamlessly insert shape adaptors into human-designed networks, we build shape adaptors on top of the same sampling functions used in the original network design. For example, in a single convolutional cell, we apply max pooling as the down-sampling layer, and the identity layer is simply an identity mapping. And in a residual cell, we use the `shortcut' $[1\times 1]$ convolutional layer as the down-sampling layer, and the weight layer stacked with multiple convolutional layers as the identity layer. In the ResNet design, we double the scaling weights in the residual cell, in order to match the same feature scale as in the original design.

Shape adaptors can also be designed in more than two branches, into a more general manner. This general design enables shape adaptors to be inserted into more complicated network architectures, such as ResNeXt \cite{xie2017aggregated} and Xception \cite{chollet2017xception}. The general formation of shape adaptors is further discussed in Appendix \ref{sec:general_shapeadaptor}.

\subsection{The Optimisation Recipe}

\paragraph{Implementations of Shape Adaptors} In practice, where we have data whose spatial dimension is an integer multiple, a different rounding method in the implementation would result in different learning dynamics. In order to enable shape adaptors perform at the highest efficiency, we propose two types of implementation aiming for specific use cases. Assuming we insert shape adaptors in every network layer, and with the spatial dimension of input data  $\mathcal{D}^{in}$, the output dimensionality of the $k^{th} (k\geq 1)$ shape adaptor module $\mathcal{D}^{(k)}$, with its corresponding reshaping factor $r^{(k)}$, is defined as:

\begin{itemize}
  \item Local type:
  \begin{equation}
  \label{eq:typelocal}
  \mathcal{D}^{(k)} = \left\lfloor \mathcal{D}^{(k-1)} \cdot r^{(k)}\right\rfloor,\quad \mathcal{D}^{(0)}=\mathcal{D}^{in}
  \end{equation}
  \item Global type:
  \begin{equation}
  \label{eq:typeglobal}
  \mathcal{D}^{(k)} = \left\lfloor \mathcal{D}^{in}\cdot \prod_{i\leq k} r^{(i)}\right\rceil
  \end{equation}
\end{itemize}
where $\lfloor\cdot\rfloor$ represents a floor function, and $\lfloor\cdot\rceil$ represents a round function.

The local type implementation corresponds to the same implementation in classical resizing layers, that is to compute the current layer dimension by reshaping the output feature dimension from the previous layer. The global type implementation is a new method introduced aiming for {\it precise resizing}: a shape adaptor reshapes input features by a holistic reshaping factor based on all previous resizing layers. This would be particularly useful when we insert a large number of resizing layers, or when we have training dataset in a small spatial dimension, and both of which could lead to a shape collapse by a local implementation (resulting in a very small shape despite having a large reshaping factor in every resizing layer). The key difference between these two types of implementations is: a local type down-sampling shape adaptor will guarantee to drop at least one spatial dimension, whilst a global type down-sampling shape adaptor can retain the spatial dimension if desired. 

For example, suppose we have training data with input spatial dimension $D^{in}=32$, optimised with a deep network composed with 20 resizing layers by the same reshaping factor $r=0.95$. The local implementation would produce an output dimension of 6, which would be much smaller than the global implementation producing an output dimension of 11. 

\paragraph{Number of Shape Adaptors} Theoretically, shape adaptors should be inserted into every network layer, to enable maximal search space and flexibility. In practice, we found that beyond a certain number of shape adaptors, performance actually began to degrade. We therefore designed a heuristic to choose an appropriate number of shape adaptor modules $N$, based on the assumption that each module contributes a roughly equal amount towards the network's overall resizing effect. Let us consider that each module resizes its input feature map in the range $(r_{min}, r_{max})$. The overall number of modules required should be sufficient to reshape the network's input dimension of $\mathcal{D}^{in}$ to a manually defined output dimension $\mathcal{D}^{last}$, by applying a sequence of reshaping operations, where each is $\sim r_{min}$. As such, the optimal number of modules can be expressed as a logarithmic function of the overall ratio between the network's input and output, based on the scale of the reshaping operation in each module:
\begin{equation}
\label{eq:number}
N = \left\lfloor \log_{1/{r_{min}}}(\mathcal{D}^{in}/\mathcal{D}^{last})\right\rfloor
\end{equation}

\paragraph{Initialisations in Shape Adaptors } As with network weights, a good initialisation for shape adaptors, i.e. the initial values for $\alpha$, is important. Again, assuming we have every shape adaptor designed in the same search space $\mathcal{R}=(r_{min}, r_{max})$ with the reshaping factor $s(\alpha)= (r_{max}-r_{min})\alpha+r_{min}$, we propose a formula to automatically compute the initialisations such that the output feature dimension of the initialised shape would map to the user-defined dimension $D^{out}$. Assuming we want to initialise the raw scaling parameters $\bar\alpha$ before sigmoid function $\alpha=\sigma(\bar\alpha)$, we need to solve the following equation:
\begin{equation}
\label{eq:init_raw}
\mathcal{D}^{in}\cdot s(\sigma(\bar{\alpha}))^N = \mathcal{D}^{out}.
\end{equation}

Suppose we use $N$ as defined in Eq. \ref{eq:number}, then Eq. \ref{eq:init_raw} is only solvable when $D^{last} \leq D^{out}$. Otherwise, we then initialise the smallest possible shape when encountering the case for $D^{last} > D^{out}$. This eventually derives the following:
\begin{equation}
  \label{eq:initialisation}
\bar{\alpha} = 
\begin{cases}
  \ln\left(-\frac{\sqrt[\leftroot{-2}\uproot{2}N]{\nicefrac{\mathcal{D}^{out}}{\mathcal{D}^{in}}} - r_{min}}{\sqrt[\leftroot{-2}\uproot{2}N]{\nicefrac{\mathcal{D}^{out}}{\mathcal{D}^{in}}} - r_{max}} +\epsilon \right) &\text{if } \mathcal{D}^{last} \leq \mathcal{D}^{out}\\
  \ln(\epsilon)  &\text{otherwise }
\end{cases}
,\quad \epsilon =10^{-4}.
\end{equation}
where $\epsilon$ is a small value to avoid encountering $\pm\infty$ values.

In practice, we need to avoid having the case when $\mathcal{D}^{last} > \mathcal{D}^{out}$, which would become a marginal point from a sigmoid function that eventually would receive a very small gradient.

\paragraph{Shape Adaptors with Memory Constraint} During experiments, we observed that shape adaptors tend to converge to a larger shape than the human designed network, which may then require very large memory. For practical applications, it is desirable to have a constrained search space for learning the optimal network shape given a user-defined memory limit. For any layer designed with down-sampling shape adaptors, the spatial dimension of which is guaranteed to be smaller than the one from the previous layers. We thus again use the final feature dimension to approximate the memory usage for the network shape.

Suppose we wish to constrain the network shape with the final feature dimension to be no greater than $D^{limit}$. We then limit the scaling factors in shape adaptors by use of a penalty value $\rho$, which is applied whenever the network's final feature dimension after the current update $D^{cout}$ is greater than the defined limit, i.e. when $D^{cout}>D^{limit}$. When this occurs, the penalty term $\rho$ is applied on every shape adaptor module, and we compute $\rho$ dynamically for every iteration so that we make sure $D^{cout} \leq D^{limit}$ in the entire training stage. The penalised scaling parameter $\alpha_\rho$ is then defined as follows,
\begin{equation}
 \alpha_\rho = \alpha \cdot \rho + \frac{r_{min}}{r_{max}-r_{min}}(\rho-1).
\end{equation} 

Then the penalised module's reshaping factor $s(\alpha_\rho)$ becomes,
\begin{equation}
s(\alpha_\rho) = (r_{max}-r_{min}) \alpha_{\rho} + r_{min} = s(\alpha)\rho.
\end{equation}

Using Eq. \ref{eq:init_raw}, we can compute $\rho$ as,
\begin{equation}
\label{eq:rho}
\rho= \sqrt[\leftroot{-3}\uproot{3}N]{\frac{\mathcal{D}^{limit}}{\mathcal{D}^{cout}}}.
\end{equation}

{\bf Iterative Optimisation Strategy\quad} To optimise a neural network equipped with shape adaptor modules, there are two sets of parameters to learn: the weight parameters $\boldsymbol{w}=\{w_i\}$, and the shape parameters $\boldsymbol{\alpha}=\{\alpha_i\}$. Unlike NAS algorithms which require optimisation of network weights and structure parameters on separate datasets, shape adaptors are optimised on the same dataset and require no re-training. 

Since the parameter space for the network shape is significantly smaller than the network weight, we update the shape parameters less frequently than the weight parameters, at a rate of once every $\alpha_s$ steps. The entire optimisation for a network equipped with shape adaptors is illustrated in Algorithm \ref{alg:shapeadaptor}.
\begin{algorithm}[ht!]
  \small
  \SetAlCapSkip{10em}
  {\bf Define:} shape adaptors: $\alpha_s, r_{min}, r_{max}, D^{last}, D^{out}, D^{limit}$\\
  {\bf Define:} network architecture $f_{\boldsymbol{\alpha, w}}$ defined with shape and network parameters\\
  {\bf Initialise:} shape parameters: $\boldsymbol{\alpha}=\{\alpha_i\}$ with Eq. \ref{eq:number}, and Eq. \ref{eq:initialisation}  \\
  {\bf Initialise:} weight parameters: $\boldsymbol{w}=\{w_i\}$ \\
  {\bf Initialise:} learning rate: $\lambda_1, \lambda_2$\\
  \While{not converged}{
  \For{each training iteration $i$}{
   $(x_{(i)}, y_{(i)}) \in (x, y)$  \hfill{\it $\triangleright$ fetch one batch of training data}\\
   \eIf{requires memory constraint}{
   {\bf Compute:} $\rho$ using Eq. \ref{eq:rho}\\
   }{{\bf Define:} $\rho=1$}
   \If{in $\alpha_s$ step}{
   {\bf Update}: $\boldsymbol{\alpha} \leftarrow \lambda_1\nabla_{\boldsymbol{\alpha}}\mathcal{L}(f_{\boldsymbol{\alpha}_\rho, \boldsymbol{w}}(x_{(i)}), y_{(i)})$  \hfill{\it $\triangleright$ update shape parameters}\\
   }
   {\bf Update}: $\boldsymbol{w} \leftarrow \lambda_2\nabla_{\boldsymbol{w}}\mathcal{L}(f_{\boldsymbol{\alpha}_\rho, \boldsymbol{w}}(x_{(i)}), y_{(i)})$  \hfill{\it $\triangleright$ update weight parameters}\\
   }
   }
   \caption{Optimisation for Shape Adaptor Networks}
   \label{alg:shapeadaptor}
 \end{algorithm}

 \section{Experiments}
 \label{sec:exp}
 In this section, we present experimental results to evaluate shape adaptors on image classification tasks. Please see the Appendix for further results, and a list of negative results for other experiments we attempted.
 
 \subsection{Experimental Setup}
 \paragraph{Datasets} We evaluate on seven different image classification datasets, with varying sizes and complexities to fully assess the robustness and generalisation of shape adaptors. These seven datasets are divided into three categories: i) small (resolution) datasets: CIFAR-10/100 \cite{krizhevsky2009learning}, SVHN \cite{goodfellow2013multi}; ii) fine-grained classification datasets: FGVC-Aircraft (Aircraft) \cite{maji13fine-grained}, CUBS-200-2011 (Birds) \cite{WahCUB_200_2011}, Stanford Cars (Cars) \cite{KrauseStarkDengFei-Fei_3DRR2013}; and iii) ImageNet \cite{deng2009imagenet}. Small datasets are in resolution $[32\times 32]$, and fine-grained classification and ImageNet datasets are in resolution $[224\times 224]$.
 
\paragraph{Baselines} We ran experiments with three widely-used networks: VGG-16 \cite{Simonyan15}, ResNet-50 \cite{he2016deep}, and MobileNetv2 \cite{sandler2018mobilenetv2}. The baseline {\it Human} represents the original human-designed networks, which require manually adjusting the number of resizing layers according to the resolution of each dataset. For smaller $[32\times 32]$ datasets, human-designed VGG-16, ResNet-50 and MobileNetv2 networks were equipped with 4, 3, 3 resizing layers respectively, and for $[224\times 224]$ datasets, all human designed networks have 5 resizing layers.
 
 \paragraph{Implementation of Shape Adaptors} For all experiments in this section, since we assume no prior knowledge of the optimal network architecture, we inserted shape adaptors uniformly into the network layers (except for the last layer). We initialised shape adaptors with $D^{last}=2, D^{out}=8$, which we found to work well across all datasets and network choices. All shape adaptors use the search space $\mathcal{R}=(0.5, 1)$ with the design in Fig. \ref{fig:structure}. We applied local type shape adaptors, to have a similar resizing effect from human-designed resizing layers, and with memory constraint on shape adaptors so that the network shape can grow no larger than the running GPU memory. We optimised shape adaptors every $\alpha_s=20$ steps for non-ImageNet datasets, and every $\alpha_s=1500$ steps for ImageNet. The full hyper-parameter choices are provided in the Appendix \ref{sec:hyper}.

 \subsection{Results on Image Classification Datasets}
 \label{sec:im_classification}
First, we compared networks built with shape adaptors to the original human-designed networks, to test whether shape adaptors can improve performances solely by finding a better network shape, without using any additional parameter space. To ensure fairness, all network weights in the human-designed and shape adaptor networks were optimised using the same hyper-parameters, optimiser, and scheduler. 

Table \ref{tab:result_full} shows the test accuracies of shape adaptor and human-designed networks, with each accuracy averaged over three individual runs. We see that in nearly all cases, shape adaptor designed networks outperformed human-designed networks by a significant margin, despite both methods using exactly the same parameter space. We also see that performance of shape adaptor designed networks are stable, with a relatively low variance across different runs. This is similar to the human-designed networks, showing stability and robustness of our method without needing the domain knowledge that is required for human-designed networks. A detailed analysis on robustness and perturbation of shape adaptors compared to other resizing modules is further discussed in Appendix \ref{sec:robustness}.

 \begin{table}[ht!]
   \vspace{1em}
   \centering
   \footnotesize
   \setlength{\tabcolsep}{0.25em}
     \begin{tabularx}{\textwidth}{l*{6}{Y}}
     \toprule
     \multicolumn{1}{c}{\multirow{2}[2]{*}{Dataset}} & \multicolumn{2}{c}{VGG-16}    & \multicolumn{2}{c}{ResNet-50} & \multicolumn{2}{c}{MobileNetv2} \\
 \cmidrule{2-7}          & {Human}& {Shape Adaptor} & {Human } & {Shape Adaptor} & {Human} & {Shape Adaptor} \\
 \midrule
   CIFAR-10 &  $94.11_{\pm 0.17}$    &   $\boldsymbol{95.35_{\pm 0.06}}$    &    $\boldsymbol{95.50_{\pm 0.09}}$    & $95.48_{\pm0.17}$     &   $93.71_{\pm 0.25}$    &$\boldsymbol{93.86_{\pm 0.23}}$   \\
     CIFAR-100 &  $75.39_{\pm 0.11}$   &  $\boldsymbol{79.16_{\pm 0.23}}$    &    $78.53_{\pm 0.11}$       &  $\boldsymbol{80.29_{\pm 0.10}}$    &   $73.80_{\pm 0.17}$      & $\boldsymbol{75.74_{\pm 0.31}}$   \\
      SVHN &  $96.26_{\pm 0.03}$   &  $\boldsymbol{96.89_{\pm 0.07}}$       &      $96.74_{\pm 0.20}$         & $\boldsymbol{96.84_{\pm 0.13}}$      & $96.50_{\pm 0.08}$  &    $\boldsymbol{96.86_{\pm 0.14}}$ \\
 \midrule
     Aircraft   &    $85.28_{\pm 0.09}$     &   $\boldsymbol{86.95_{\pm 0.29}}$      &      $81.57_{\pm 0.51}$    &       $\boldsymbol{85.60_{\pm 0.32}}$         &   $77.64_{\pm 0.23}$  &  $\boldsymbol{83.00_{\pm 0.30}}$   \\
     Birds  &     $73.37_{\pm0.35}$          &     $\boldsymbol{74.86_{\pm 0.50}}$         &  $68.62_{\pm 0.10}$     &    $\boldsymbol{71.02_{\pm 0.48}}$         &  $60.37_{\pm 1.12}$  &  $\boldsymbol{68.53_{\pm 0.21}}$   \\
     Cars  &     $89.30_{\pm 0.21}$        &  $\boldsymbol{90.13_{\pm 0.11}}$    &   $87.23_{\pm 0.48}$      &    $\boldsymbol{89.67_{\pm 0.20}}$     &     $80.86_{\pm 0.13}$      &     $\boldsymbol{84.62_{\pm 0.38}}$    \\
     \midrule
     ImageNet  &    $\boldsymbol{73.92_{\pm 0.12}}$   &  $73.53\pm_{0.09}$   &   $77.18_{\pm 0.04}$    &   $\boldsymbol{78.74_{\pm 0.12}}$    &   $71.72_{\pm 0.02}$    &   $\boldsymbol{73.32_{\pm 0.07}}$    \\
     \bottomrule
     \end{tabularx}%
       \caption{Top-1 test accuracies on different datasets for networks equipped with human-designed resizing layers and with shape adaptors. We present the results with the range of three independent runs. Best results are in bold.}
   \label{tab:result_full}
 \end{table}

Note that shape adaptors presented here are optimised purely to achieve an optimal performance, in a defined representation space, without considering the expense of memory consumption. However, we may also design memory-efficient shape adaptors for network compression which we will present in Section \ref{sec:autosc}.

\subsection{Ablative Analysis \& Visualisations}
In this section, we perform an ablative analysis on CIFAR-100 and Aircraft datasets to understand the behaviour of shape adaptors with respect to the number of shape adaptors, and shape adaptor initialisation. We observed that conclusions are consistent across different networks, thus we perform experiments in two networks only: VGG-16 and MobileNetv2. All results are averaged over two independent runs.

\subsubsection{Number of Shape Adaptors}
\label{sec:ablative_number}

We first evaluate the performance by varying different number of shape adaptors used in the network, whilst fixing all other hyper-parameters used in Section \ref{sec:im_classification}. In Table \ref{tab:ablative_number}, we show that the performance of shape adaptor networks is consistent across the number of shape adaptors used. Notably, performance is always better than networks with human-designed resizing layers, regardless of the number of shape adaptors used. This again shows the ability of shape adaptors to automatically learn optimal shapes without requiring domain knowledge. The optimal number of shape adaptor modules given by our heuristic in Eq. \ref{eq:number} is highlighted in teal, and we can therefore see that this is a good approximation to the optimal number of modules.

\begin{table}[t!]
  \centering
  \scriptsize
  \setlength{\tabcolsep}{0.05em}
  \begin{subtable}{0.48\linewidth}
    \centering
    \begin{tabularx}{\textwidth}{lc*{5}{Y}}
    \toprule
    \multirow{2}[2]{*}{CIFAR-100}  & \multirow{2}[2]{*}{Human} & \multicolumn{5}{c}{Shape Adaptor (with number of)} \\
        \cmidrule{3-7}
        &       & 3     & \cellcolor{lightteal} 4     &   5 & 6     & 8      \\
    \midrule
    VGG-16 & $75.39$       &  79.03     &  $\boldsymbol{79.16}$     &  $78.56$   &  $78.43$     &   $78.16$     \\
    MobileNetv2 & 73.80      &    $75.39$     &  $\boldsymbol{75.74}$    &  $75.22$     &   $74.92$       &  $74.86$      \\
    \bottomrule
  \end{tabularx}
    \end{subtable}%
    \quad\,
    \begin{subtable}{0.48\linewidth}
    \centering
    \begin{tabularx}{\textwidth}{lc*{5}{Y}}
      \toprule
    \multirow{2}[2]{*}{Aircraft}   & \multirow{2}[2]{*}{Human} & \multicolumn{5}{c}{Shape Adaptor (with number of)} \\
        \cmidrule{3-7}
          &      &  5    &   6 \cellcolor{lightteal}    &   7   & 8   & 10  \\
    \midrule
    VGG-16 &    $85.28$  &   84.80   &   $\boldsymbol{86.95}$      &   86.44    &  86.72  &85.76   \\
    MobileNetv2 &    $77.64$  &  81.12    &   83.00     &  $\boldsymbol{83.02}$  &     82.43  & 80.36    \\
    \bottomrule
    \end{tabularx}%
    \end{subtable}
    \caption{Test accuracies of VGG-16 on CIFAR-100 and MobileNetv2 on Aircraft, when different numbers of shape adaptors are used. Best results are in bold. The number produced in Eq. \ref{eq:number} is highlighted in teal.}
  \label{tab:ablative_number}
\end{table}

\begin{figure}[ht!]
  \setlength{\tabcolsep}{0.2em}
  \footnotesize
  \begin{tabularx}{\textwidth}{*{8}{Y}}
    \multicolumn{4}{c}{VGG-16 on CIFAR-100}  &
    \multicolumn{4}{c}{MobileNetv2 on Aircraft} \\
    \includegraphics[width=\linewidth, height=1.2\linewidth]{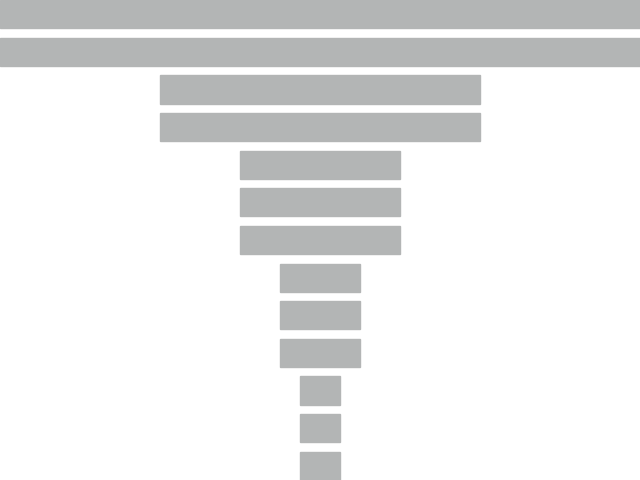} &
    \includegraphics[width=\linewidth, height=1.2\linewidth]{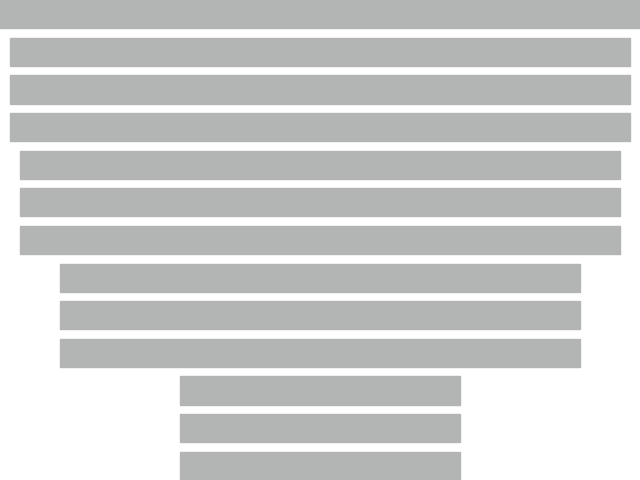} &
    \includegraphics[width=\linewidth, height=1.2\linewidth]{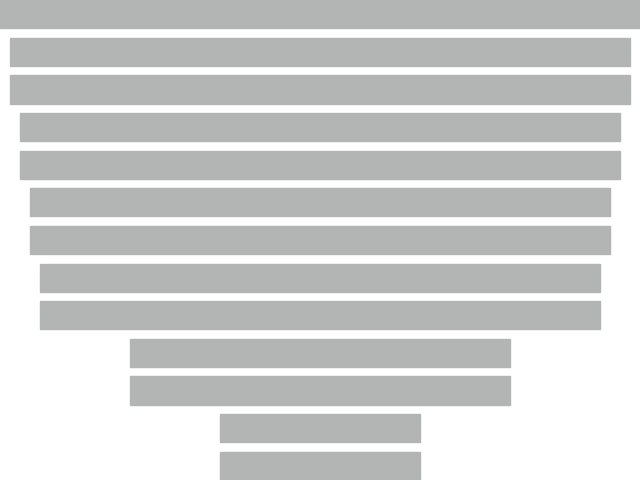} &
    \includegraphics[width=\linewidth, height=1.2\linewidth]{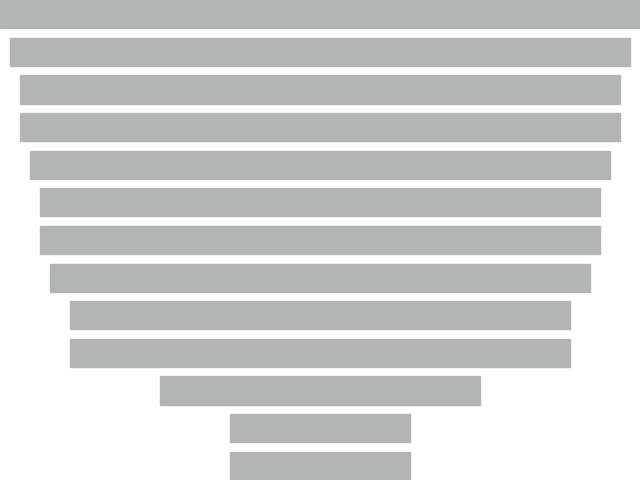} &
    \includegraphics[width=\linewidth, height=1.2\linewidth]{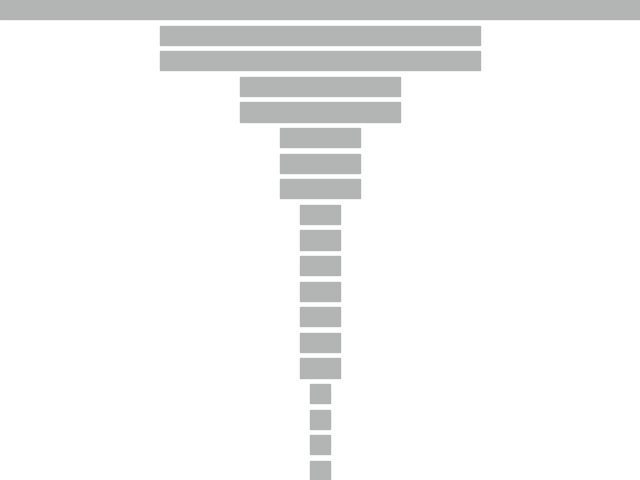} &
    \includegraphics[width=\linewidth, height=1.2\linewidth]{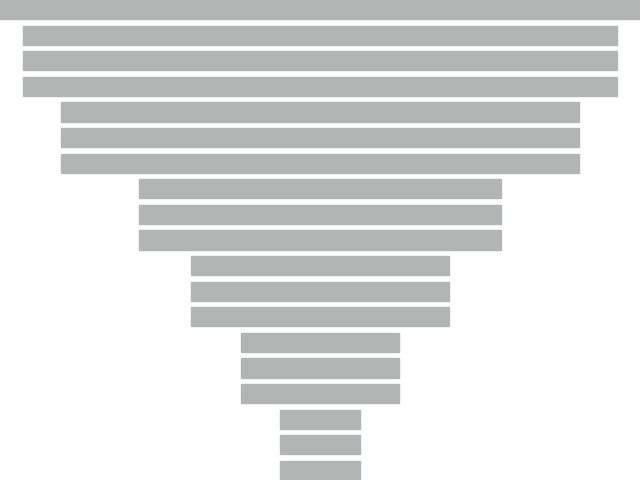} &
    \includegraphics[width=\linewidth, height=1.2\linewidth]{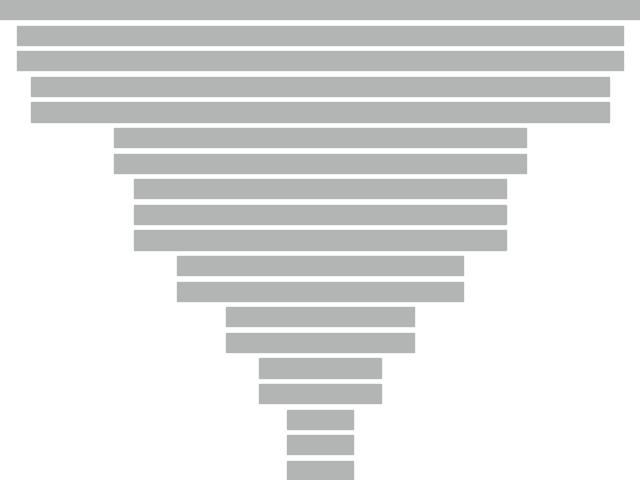} &
    \includegraphics[width=\linewidth, height=1.2\linewidth]{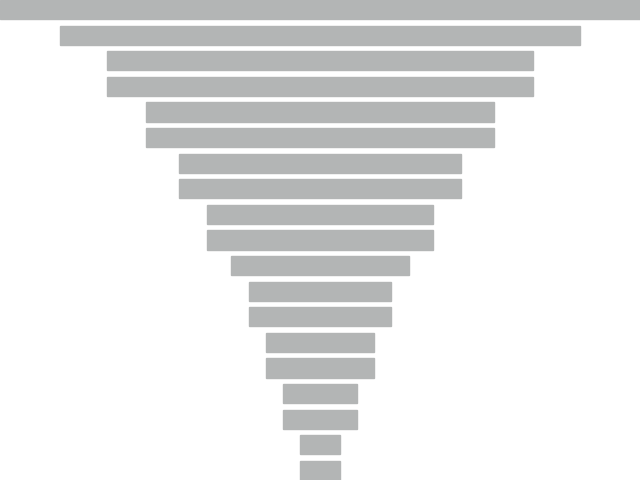} \\
    4 & 4 & 6 & 8 & 5 & 6 & 8 & 10 \\
     {\it Human Designed}  & \multicolumn{3}{c}{\it Shape Adaptor Designed} & {\it Human Designed}  & \multicolumn{3}{c}{\it Shape Adaptor Designed} \\
  \end{tabularx}
  \caption{Visualisation of human-designed and shape adaptor designed network shapes. The number on the second row represents the number of resizing layers (or shape adaptors) applied in the network. }
  \label{fig:vis_number}
\end{figure}

In Figure \ref{fig:vis_number}, we present visualisations of network shapes in human-designed and shape adaptor designed networks. We can see that the network shapes designed by our shape adaptors are visually similar when different numbers of shape adaptor modules are used. In Aircraft dataset, we see a narrower shape with MobileNetv2 due to inserting an excessive number of 10 shape adaptors, which eventually converged to a local minima and lead to a degraded performance.   

\subsubsection{Initialisations in Shape Adaptors}
\label{sec:ablative_init}

Here, we evaluate the robustness of shape adaptors by varying initialisation of $\alpha$. Initialisation with a ``wide" shape (large $\alpha$) causes high memory consumption and a longer training time, whereas initialisation with a ``narrow" shape (small $\alpha$) results in weaker gradient signals and a more likely convergence to a non-optimal local minima. In Table \ref{tab:ablative_init}, we can see that the performance is again consistently better than the human-designed architecture, across all tested initialisations. The initialisation in shape adaptor modules given by our formula Eq. \ref{eq:initialisation} is highlighted in teal.

\begin{table}[ht!]
  \centering
  \scriptsize
  \setlength{\tabcolsep}{0.05em}
  \begin{subtable}{0.48\linewidth}
    \centering
    \begin{tabularx}{\textwidth}{lc*{4}{Y}}
    \toprule
    \multirow{2}[2]{*}{CIFAR-100}   & \multirow{2}[2]{*}{Human} & \multicolumn{4}{c}{Shape Adaptor (with $s(\alpha)$ initialised)} \\
        \cmidrule{3-6}
          &        &  0.60      & \cellcolor{lightteal} 0.70      & 0.80   & 0.90\\
    \midrule
    VGG-16 & $75.39$     &    $\boldsymbol{79.21}$    &  $79.16$     &    78.79   &   78.53    \\
    MobileNetv2 &   73.80    &   75.16     &    $\boldsymbol{75.74}$     &  74.89      &   74.74    \\
    \bottomrule
  \end{tabularx}
    \end{subtable}%
    \quad\,
    \begin{subtable}{0.48\linewidth}
    \centering
    \begin{tabularx}{\textwidth}{lc*{4}{Y}}
      \toprule
    \multirow{2}[2]{*}{Aircraft}  & \multirow{2}[2]{*}{Human} & \multicolumn{4}{c}{Shape Adaptor (with $s(\alpha)$ initialised)} \\
          \cmidrule{3-6}
          &       &  0.52      & \cellcolor{lightteal} 0.58      & 0.62   & 0.68 \\
    \midrule
    VGG-16 &   85.28   &   83.90    &   $\boldsymbol{86.95}$       &   86.36   & 86.54 \\
    MobileNetv2 &   $77.64$     &  79.64     &   $\boldsymbol{83.00}$     & 82.51  & 81.56   \\
    \bottomrule
    \end{tabularx}
    \end{subtable}
    \caption{Test accuracies of VGG-16 on CIFAR-100 and MobileNetv2 on Aircraft datasets in shape adaptors with different initialisations. Best results are in bold. The initialisation produced in Eq. \ref{eq:initialisation} is highlighted in teal.}
  \label{tab:ablative_init}
\end{table}

\begin{figure}[ht!]
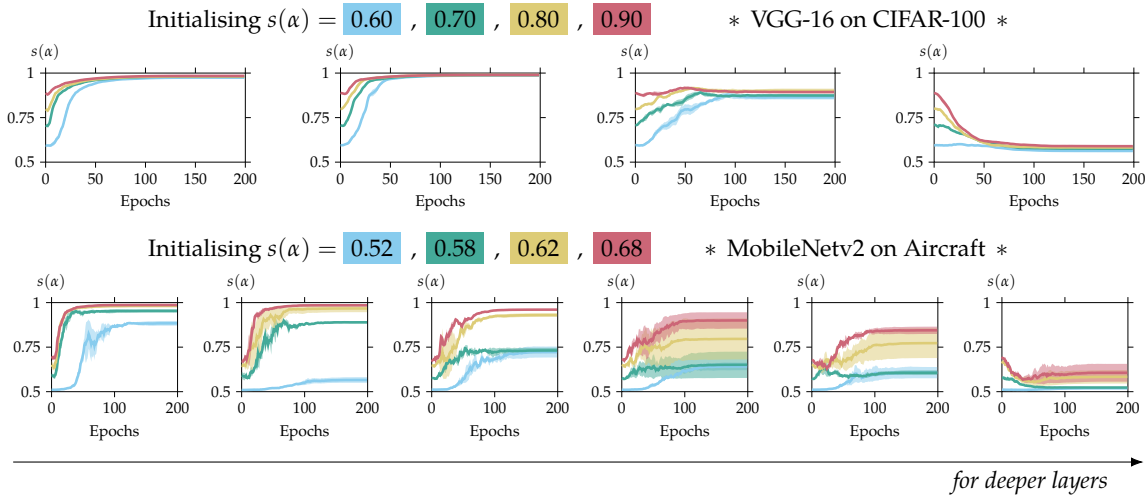

   \footnotesize
   \renewcommand{\arraystretch}{1.4}
   \setlength{\tabcolsep}{0.0em}
   \begin{tabularx}{\textwidth}{*{4}{Y}}
   \multicolumn{4}{c}{Initialising $s(\alpha) = $ \colorbox{cyan}{0.60} \,,\, \colorbox{teal}{0.70} \,,\, \colorbox{sand}{0.80} \,,\, \colorbox{rose}{0.90} \qquad\quad $\ast$\, VGG-16 on CIFAR-100 \,$\ast$} \\
      \multicolumn{4}{c}{\scalebox{0.49}{\large \input{image/vgg_cifar100_0.tex}\qquad \input{image/vgg_cifar100_1.tex}\qquad \input{image/vgg_cifar100_2.tex} \qquad \input{image/vgg_cifar100_3.tex}} }\\
     \multicolumn{4}{c}{Initialising $s(\alpha) = $ \colorbox{cyan}{0.52} \,,\, \colorbox{teal}{0.58} \,,\, \colorbox{sand}{0.62} \,,\, \colorbox{rose}{0.68}  \qquad $\ast$\, MobileNetv2 on Aircraft \,$\ast$} \\
     \multicolumn{4}{c}{\scalebox{0.49}{\large \input{image/mobilenetv2_aircraft_0.tex}\input{image/mobilenetv2_aircraft_1.tex}\input{image/mobilenetv2_aircraft_2.tex}\input{image/mobilenetv2_aircraft_3.tex}\input{image/mobilenetv2_aircraft_4.tex}\input{image/mobilenetv2_aircraft_5.tex}}}\\
     \multicolumn{4}{c}{\begin{tikzpicture}
     \draw [-Latex] (0,0) -- (0.97\textwidth,0) node [pos=0.9, below] {\it for deeper layers};
     \end{tikzpicture}}
   \end{tabularx}
  \caption{Visualisation of learning dynamics for every shape adaptor module across the entire training stage.}
  \label{fig:vis_init}
\end{figure}

In Figure \ref{fig:vis_init}, we present the learning dynamics for each shape adaptor module across the entire training stage. We can observe that shape adaptors are learning in an almost identical pattern across different initialisations in the CIFAR-100 dataset, with nearly no variance. For the larger resolution Aircraft dataset, different initialised shape adaptors converged to a different local minimum. They still follow a general trend, for which the reshaping factor of a shape adaptor inserted in the deeper layers would converge into a smaller scale.

% This is precise version based on tikz.

\subsection{A Detailed Study on Neural Shape Learning}
In this section, we propose a study to analyse different neural shape learning strategies, and the transferability of learned shapes. Likewise, all results are averaged over two independent runs.

We evaluate different neural shape learning strategies by running shape adaptors in three different versions. {\it Standard}: the standard implementation from previous sections; {\it Fix (Final)}: a network retrained with a fixed optimal shape obtained from shape adaptors; and {\it Fix (Large)}: a network retrained with a fixed largest possible shape in the current running GPU memory. The Fix (Final) baseline is designed to align with the training strategy from NAS algorithms \cite{liu2018darts,zoph2017neural,pmlr-v80-pham18a}. The Fix (Large) baseline is to test whether naively increasing network computational cost can give improved performance.

\begin{table}[ht!]
  \centering
  \scriptsize
  \setlength{\tabcolsep}{0.15em}
  \begin{subtable}{0.48\linewidth}
    \centering
    \begin{tabularx}{\textwidth}{lc*{3}{Y}}
    \toprule
  \multirow{2}[2]{*}{CIFAR-100}   & \multirow{2}[2]{*}{Human} & \multicolumn{3}{c}{Shape Adaptors} \\
      \cmidrule{3-5}
        &        &  Standard    & \makecell{Fix (Final)}     & \makecell{Fix (Large)} \\
  \midrule
  \multirow{2}[0]{*}{VGG-16} & $75.39$     &   79.16     &  78.62    &   78.51    \\
   &   314M   & 5.21G  & 5.21G  & 9.46G \\
   \midrule
   \multirow{2}[0]{*}{MobileNetv2} &   73.80    &   75.74     &  75.54   & 75.46      \\
  &   94.7M   & 923M  &  923M  & 1.35G  \\
  \bottomrule
  \end{tabularx}
  \end{subtable}%
  \qquad
  \begin{subtable}{0.48\linewidth}
  \centering
  \begin{tabularx}{\textwidth}{lc*{3}{Y}}
    \toprule
  \multirow{2}[2]{*}{Aircraft}  & \multirow{2}[2]{*}{Human} & \multicolumn{3}{c}{Shape Adaptors} \\
        \cmidrule{3-5}
        &       &  Standard     &   \makecell{Fix (Final)}     &  \makecell{Fix (Large)} \\
  \midrule
  \multirow{2}[0]{*}{VGG-16} &   85.28   &     86.95      &   86.27   &  84.49  \\
  &   15.4G   & 50.9G  & 50.9G   & 97.2G  \\
  \midrule
  \multirow{2}[0]{*}{MobileNetv2} &   $77.64$      &   $83.00$     &   82.26 &   81.18  \\
      &   326M   & 9.01G  &  9.01G   &  12.0G \\
  \bottomrule
  \end{tabularx}
  \end{subtable}
  \caption{Test accuracies and computational cost (MACs, the number of multiply-adds) on CIFAR-100 and Aircraft datasets trained with different shape learning strategies.}
\label{tab:ablative_shapelearning}
\end{table}

In Table \ref{tab:ablative_shapelearning}, we can observe that our standard version achieves the best performance among all shape learning strategies. In addition, we found that just having a large network would not guarantee an improved performance (VGG-16 on Aircraft). This validates that shape adaptors are truly learning the optimal shape, rather than naively increasing computational cost. Finally, we can see that our original shape learning strategy without re-training performs better than a NAS-like two-stage training strategy, which we assume is mainly due to dynamically updating of network shape helping to learn spatial-invariant features. 

In order to further understand how network performance is correlated with different network shapes, we ran a large-scale experiment by training 200 VGG-16 networks with randomly generated shapes.

\begin{minipage}{\textwidth}
  \begin{adjustbox}{valign=b, minipage={.63\textwidth}}
  \notsotiny
  \renewcommand{\arraystretch}{1.6}
  \setlength{\tabcolsep}{0.1em}
  \begin{tabularx}{\textwidth}{*{4}{Y}}
    \multicolumn{4}{c}{VGG-16 on CIFAR-100}\\
    \includegraphics[width=0.9\linewidth, height=1.1\linewidth]{image/vgg_cifar100_human} &
    \includegraphics[width=0.9\linewidth, height=1.1\linewidth]{image/vgg_cifar100_sanum4} &
    \includegraphics[width=0.9\linewidth, height=1.1\linewidth]{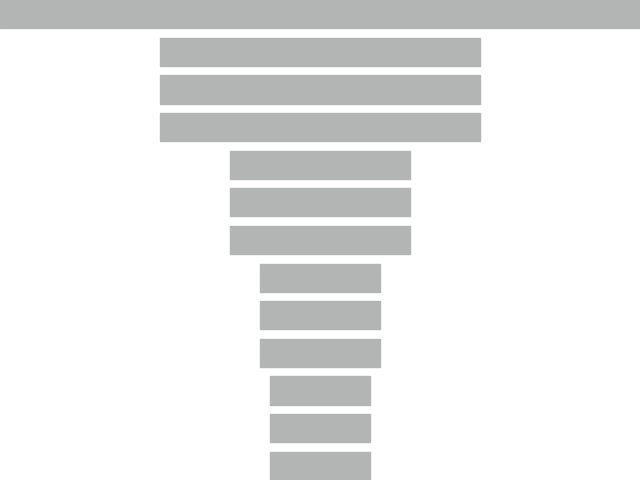} &
    \includegraphics[width=0.9\linewidth, height=1.1\linewidth]{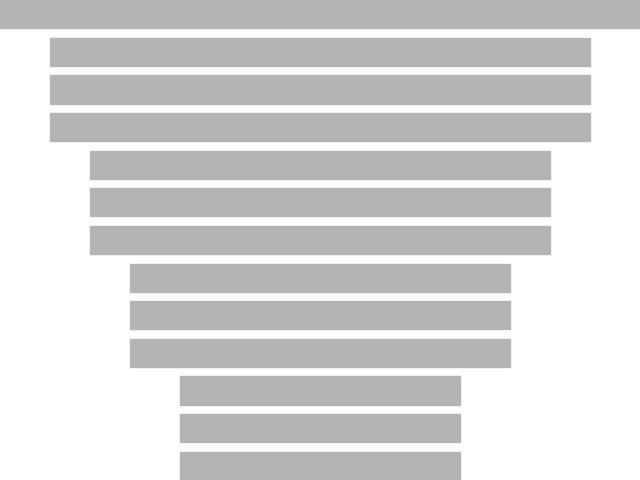}  \\
    Acc: 75.39 & Acc: 79.16 & Acc: 73.86 & Acc: 79.12  \\\relax
    [0.50, 0.50, 0.50, 0.50]  & [0.98, 0.98, 0.88, 0.57]  &
    [0.50, 0.60, 0.75, 0.96]  & [0.86, 0.85, 0.84, 0.74]  \\
     {\it    \tikzbullet{orange}{orange}  Human Designed}  & {\it  \tikzbullet{red}{red} Shape Adaptors} & {\it  \tikzbullet{black}{white} Worst Random Search}  & {\it \tikzbullet{black}{white} Best Random Search}   \\
  \end{tabularx}
  \end{adjustbox}
  \begin{adjustbox}{valign=b,minipage={.37\textwidth}}
    \scalebox{0.72}{\input{image/random_search.tex}}
  \end{adjustbox}
  \captionof{figure}{Visualisation and test accuracies of VGG-16 on CIFAR-100 in 200 randomly generated shapes. The second row represents the precise reshaping factor in each resizing layer}
  \label{fig:random}
  \vspace{2em}
\end{minipage}

In Fig. \ref{fig:random}, we visualise the randomly generated network shapes with the best and the worst performance, and compare them to the network shapes in human designed and shape adaptor networks. 

First, we can see that the best randomly searched shape obtains a very similar performance as well as a similar structure of shape compared to the ones learned from shape adaptors. Second, the reshaping factors in the worst searched shape are arranged from small to large, which is the direct opposite trend to the reshaping factors automatically learned by our shape adaptors. Third, human-designed networks are typically under-sized, and just by increasing network memory cost is not able to guarantee an improved performance. Finally, we can see a clear correlation between memory cost and performance, where a higher memory cost typically increases performance. However, this correlation ceases after 5G of memory consumption, after which point we see no improved performance. Interestingly, the memory cost of shape adaptors lies just on the edge of this point, which again shows the shape adaptor's ability to learn optimal design.

\section{Other Applications}
In this section, we present two additional applications of shape adaptors: Automated Shape Compression (AutoSC) and Automated Transfer Learning (AutoTL).

\subsection{Automated Shape Compression}
\label{sec:autosc}

In previous sections, we have shown that shape adaptors are able to improve performance by finding the optimal network shapes, but with a cost of a huge memory requirement of the learned network. In AutoSC, we show that shape adaptors can also achieve strong results, when automatically finding optimal memory-bounded network shapes based on an initial human design. Instead of the original implementation of shape adaptors where these are assumed to be the only resizing layers in the network, with AutoSC we attach down-sampling shape adaptors only on top of the non-resized layers of the human-designed architecture, whilst keeping the original human-designed resizing layers unchanged. Here, we insert global type shape adaptors, to be initialised so that the network shape is identical to the human-designed architecture, and thus the down-sampling shape adaptors can only learn to compress the network shape. This guarantees that the learned shape requires no more memory than the human-designed shape.

\begin{table}[ht!]
  \vspace{0.8em}
  \footnotesize
  \setlength{\tabcolsep}{0.3em}
  \centering
  \begin{subtable}[t]{0.43\linewidth}
    \centering
    \begin{tabular}{lcccc}
      \toprule
      {\bf 200/300M} MobileNetv2 & Params& MACs  & \makecell{Acc.}\\
      \midrule
      Human \(0.75\times\) & 2.6M  & 233M  &69.8\\
      AutoSC \(0.85\times\) & 2.9M  & 262M & 70.7\\
      \midrule
      Human \(1.0\times\)  & 3.5M & 330M  &71.8\\
    AutoSC \(1.1\times\) & 4.0M & 324M& 72.3 \\
    \bottomrule
  \end{tabular}
  \caption{\footnotesize Results on ImageNet}
\end{subtable}
\qquad 
  \begin{subtable}[t]{0.5\linewidth}
  \centering
    \begin{tabularx}{\textwidth}{l*{3}{Y}}
    \toprule
     {\bf Plain} MobileNetv2 & Params& MACs  & \makecell{Acc.}\\
      \midrule
      Human \(1.0\times\) & 2.3M  & 94.7M  & 73.80\\
      AutoSC \(1.0\times\)  & 2.3M  & 91.5M & 74.81\\
      \midrule
      Human \(1.0\times\)  & 2.3M & 330M  &77.64\\
    AutoSC \(1.0\times\) & 2.3M & 326M& 78.95\\
    \bottomrule
    \end{tabularx}
    \caption{\footnotesize\centering Results on CIFAR-100 (up) and Aircraft (down)}
    \end{subtable}
   \caption{Test accuracies for AutoSC and human-designed MobileNetv2 on CIFAR-100, Aircraft, and ImageNet. $\times$ represents the applied width multiplier.}
  \label{tab:result_autosc}
\end{table}

In Table \ref{tab:result_autosc}, we present AutoSC built on MobileNetv2, an efficient network design for mobile applications. We evaluate AutoSC on three datasets: CIFAR-100, Aircraft and ImageNet. During training of MobileNetv2, we initialised a small width multiplier on the network's channel dimension to slightly increase the parameter space (if applicable). By doing this, we ensure that this ``wider" network after compression would have a similar memory consumption as the human-designed MobileNetv2, for a fair comparison. In all three datasets, we can observe that shape adaptors are able to improve performance, despite having similar memory consumption compared to human-designed networks.

\subsection{Automated Transfer Learning}
In this section, we present how shape adaptors can be used to perform transfer learning in an architectural level. In AutoTL, we directly replace the human-designed resizing layers with shape adaptors, and initialise them with the reshaping factors designed in the original human-defined architecture, to match the spatial dimension of each pre-trained network layer. During fine-tuning, the network is then fine-tuning with network weights along with network shapes, thus improving upon the standard fine-tuning in a more flexible manner.

We follow the same setting as in PackNet \cite{mallya2018piggyback} and Piggyback \cite{mallya2018piggyback}, evaluating on 5 fine-grained classification datasets across very different domains. For all tasks, we use input images of resolution $[224\times 224]$, and evaluate them on an ImageNet pre-trained ResNet-50.

\begin{table}[ht!]
  \centering
  \small
    \begin{tabularx}{\textwidth}{l*{5}{Y}}
      \toprule
      & Birds \cite{WahCUB_200_2011} & Cars \cite{KrauseStarkDengFei-Fei_3DRR2013} & Flowers \cite{nilsback2008automated} & WikiArt \cite{Saleh_Elgammal_2016} & Sketches \cite{eitz2012humans}\\
      \midrule
    PackNet \cite{mallya2018packnet} &  80.41   &   86.11   &   93.04   &   69.40  &76.17 \\
    PiggyBack \cite{mallya2018piggyback} &  81.59    &   89.62    &   94.77   &   71.33  & 79.91 \\
    NetTailor \cite{morgado2019nettailor} &  82.52     &   90.56    &   95.79    &    72.98   &  80.48\\
    \midrule
    Fine-tune \cite{guo2019spottune} & 81.86   &   89.74    &   93.67     &    75.60  &  79.58 \\
    SpotTune \cite{guo2019spottune} &  84.03     &   92.40    &   {\bf 96.34}    &    75.77   &  80.20\\
    AutoTL &   {\bf 84.29}    &  {\bf 93.66}      &  96.22   &    {\bf 77.47}   & {\bf 80.74}  \\
    \bottomrule
    \end{tabularx}%
    \caption{Test accuracies of transfer Learning methods built on ResNet-50 on fine-grained datasets. Best results are in bold. }
  \label{tab:autotl}%
\end{table}

The results for AutoTL and other state-of-the-art transfer learning methods are listed in Table \ref{tab:autotl}, for which we outperform 4 out of 5 datasets. The most related methods to our approach are standard fine-tuning and SpotTune \cite{guo2019spottune}, which optimise the entire network parameters for each dataset. Other approaches like PackNet \cite{mallya2018packnet}, Piggyback \cite{mallya2018piggyback}, and NetTailor \cite{morgado2019nettailor} focus on efficient transfer learning by updating few task-specific weights. We design AutoTL with standard fine-tuning, as the simplest setting to show the effectiveness of shape adaptors. In practice, AutoTL can be further improved, and integrated into other efficient transfer learning techniques.

\section{Conclusions \& Future Directions}
In this paper, we present shape adaptor, a learnable resizing module to enhance existing neural networks with task-specific network shapes. With shape adaptors, the learned network shapes can further improve performances compared to human-designed architectures, without requiring in increase in parameter space. We show that shape adaptors are robust to hyper-parameters, and typically learn very similar network shapes, regardless of the number of shape adaptor modules used. In addition, we show that shape adaptors can also be easily incorporated into other applications, such as network compression and transfer learning.

In future work, we will investigate shape adaptors in a multi-branch design, where the formulation provided in this paper extended to integrating more than two resizing layers in each shape adaptor module. Due to the success of shape adaptors in the other applications we have presented in this paper, we will also study use of shape adaptors for more applications, such as neural architecture search, and multi-task learning.

\section*{Acknowledgements}
We would like to thank the helpful discussions from Linxi Fan and Bingyu Zhang.

\bibliography{egbib}
\bibliographystyle{plain}

\newpage
\appendix

\section{General Formation of Shape Adaptors}
\label{sec:general_shapeadaptor}
Shape adaptors can be extended into a multi-branch design, into a more general manner. Each shape adaptor module is then composed of $K\geq 2$ resizing layers $F_{i=1:K}$, with fixed reshaping factors $r_{i=1:K}>0$, and the corresponding learnable scaling weight parameters $\alpha_{i=1:K}\in(0,1)$. In such design, shape adaptor modules can not only learn the optimal network shape, but also the optimal operations contributing to the learned shape.

We first define a set of reshaping factors $r_i$, and scaling weights $\alpha_i$ in resizing layers $F_i$:
\begin{equation}
\boldsymbol{r}=\left\{{r}_{i=1:K}\,\middle|\, r_i >0,\, \exists m,n: r_m
\neq r_n \right\}, \quad \text{and}\quad\boldsymbol{\alpha}=\left\{{\alpha}_{i=1:K} \, \middle|\, \sum_{i=1}^K \alpha_i=1,\, \alpha_i> 0\right\}.
\end{equation}
 
Then, we design the module's reshaping factor which is mapped from scaling weights $\boldsymbol{\alpha}\to s(\boldsymbol{\alpha})$, lying in the defined search space interval $\mathcal{R}=(\min(\boldsymbol{r}),\, \max(\boldsymbol{r}))$. 

The general system for a shape adaptor module is formulated as follows,
\begin{equation}
{\tt ShapeAdaptor}(x, \boldsymbol{\alpha}, \mathbf{r}) = \sum_{i=1}^K \alpha_i \cdot G\left(F_i\left(x, r_i\right),~ \frac{s(\boldsymbol{\alpha})}{r_i}\right),
\end{equation}
with $s(\boldsymbol{\alpha})$ satisfies
\begin{equation}
s(\boldsymbol{\alpha})_{\alpha_k\to 1} = r_k, \quad \text{and}\quad s(\boldsymbol{\alpha}) \mapsto \mathcal{R}.
\end{equation}

The weighted generalised mean: 
\begin{equation}
s_0(\boldsymbol{\alpha}) = \prod_{i=1}^K r_i^{\alpha_i},\quad \text{and}\quad s_p(\boldsymbol{\alpha}) = \left(\sum_{i=1}^K\alpha_i r_i^p\right)^{1/p},\, p\neq 0
\end{equation}
are examples of suitable reshaping function design.

The general design for multi-branch shape adaptors can be inserted into more complicated networks architectures, such as ResNeXt \cite{xie2017aggregated} and Xception \cite{chollet2017xception}. It can also be seen as a direct enhancement to spatial pyramid pooling \cite{he2015spatial,chen2017deeplab}, and U-Net \cite{ronneberger2015u}, to enable them to propagate context information from various, rather than the same, feature dimensions.

\newpage
\section{The Complete Hyper-Parameter Table}
\label{sec:hyper}

In this section, for reproducibility, we present a detailed list of hyper-parameter choices, across all networks and datasets evaluated in Table 1. $A$ represents network shape parameters in shape adaptors, and $W$ represents network weight parameters.

\begin{table}[ht!]
  \centering
  \scriptsize
  \renewcommand{\arraystretch}{2}
  \setlength{\tabcolsep}{0.0em}
  \begin{tabularx}{\textwidth}{l*{9}{Y}}
    \toprule
     & \multicolumn{3}{c}{Small Datasets: $[32 \times 32]$} & \multicolumn{3}{c}{Fine-Grained Datasets: $[224 \times 224]$} & \multicolumn{3}{c}{ImageNet: $[224 \times 224]$} \\
\cmidrule{2-10}          & VGG-16 & ResNet-50 & MobileNetv2    & VGG-16 & ResNet-50 & MobileNetv2  & VGG-16 & ResNet-50 & MobileNetv2 \\
   \midrule
   $A$ - Learning Rate & \multicolumn{3}{c}{0.1} & \multicolumn{3}{c}{0.1} & \multicolumn{3}{c}{0.1} \\
    \midrule
    $A$  - Optimiser & \multicolumn{3}{c}{SGD with 0.9 momentum} & \multicolumn{3}{c}{SGD with 0.9 momentum} & \multicolumn{3}{c}{SGD with 0.9 momentum} \\
    \midrule
    $A$  - Scheduler & \multicolumn{3}{c}{Cosine Annealing} & \multicolumn{3}{c}{Cosine Annealing} & \multicolumn{3}{c}{Cosine Annealing} \\
    \midrule
    $A$  - Update Step & \multicolumn{3}{c}{20} & \multicolumn{3}{c}{20} & \multicolumn{3}{c}{1500} \\
    \midrule
    $A$  - Number & \multicolumn{9}{c}{Eq. 2: $\log_2(D^{in}/2)$ (4 for $[32\times 32]$ images, 6 for $[224\times 224]$ images)} \\
    \midrule
    $A$  - Initialisation & \multicolumn{9}{c}{Eq. 4 with $D^{out}=8$}\\
    \midrule
    $A$  - Location & \multicolumn{9}{c}{Uniformly distributed (across all layers except for the last layer)} \\
    \midrule
    $A$  - Search Space & \multicolumn{9}{c}{ $(0.5, 1.0)$ (for every shape adaptor module)} \\
    \midrule
    $W$ - Learning Rate & \multicolumn{3}{c}{0.1} & \multicolumn{3}{c}{0.01} & \multicolumn{1}{c}{0.1} & \multicolumn{1}{c}{0.1} & \multicolumn{1}{c}{0.05} \\
    \midrule
    $W$ - Optimiser & \multicolumn{3}{c}{SGD with 0.9 momentum} & \multicolumn{3}{c}{SGD with 0.9 momentum} & \multicolumn{3}{c}{SGD with 0.9 momentum} \\
    \midrule
    $W$ - Weight Decay & \multicolumn{1}{c}{$5\cdot 10^{-4}$}  & \multicolumn{1}{c}{$5\cdot 10^{-4}$}  & \multicolumn{1}{c}{$4\cdot 10^{-5}$} & \multicolumn{1}{c}{$5\cdot 10^{-4}$} & \multicolumn{1}{c}{$5\cdot 10^{-4}$} & \multicolumn{1}{c}{$4\cdot 10^{-5}$} & \multicolumn{1}{c}{$5\cdot 10^{-4}$} & \multicolumn{1}{c}{$5\cdot 10^{-4}$} & \multicolumn{1}{c}{$4\cdot 10^{-5}$} \\
     \midrule
    $W$ - Scheduler & \multicolumn{3}{c}{Cosine Annealing} & \multicolumn{3}{c}{Cosine Annealing} & \multicolumn{3}{c}{Cosine Annealing} \\
    \midrule
    Batch Size & \multicolumn{3}{c}{128} & \multicolumn{3}{c}{8} & \multicolumn{3}{c}{32 (per GPU) for 8 GPUs} \\
    \midrule
    Epochs & \multicolumn{3}{c}{200} & \multicolumn{3}{c}{200} & \multicolumn{3}{c}{120} \\
    \bottomrule
    \end{tabularx}%
    \caption{The complete hyper-parameter applied to reproduce Table 1.}
  \label{tab:hyper}%
\end{table}

\newpage

\section{Corruptions and Perturbations Analysis} 
\label{sec:robustness}

In this section, we evaluate the model robustness and uncertainty estimates in networks equipped with shape adaptors, compared with other types of resizing modules.

\paragraph{Metrics} We apply two metrics with respect to corruption and perturbation robustness evaluation respectively, introduced in \cite{hendrycks2018benchmarking}. For corruption analysis, we evaluate with {\it mean Corruption Error (mean CE)}, which computes an average classification error on a corrupted dataset, composed by corrupting the original dataset with 15 corruption types, and each with additional 5 severity levels. For perturbation analysis, each data in a perturbed dataset becomes a video, to measure prediction stability. We then evaluate by measuring whether video frames prediction match, which we call flip probability. We evaluate with 10 perturbation types, and the mean across these is {\it mean Flip Rate (mean FR)}.

We apply corruption analysis on CIFAR-10 and CIFAR-100 dataset, which give the corrupted CIFAR-10-C and CIFAR-100-C datasets respectively. We apply perturbation analysis on CIFAR-10 only, for perturbed CIFAR-10-P dataset, with the highest difficulty level 3. All analyses are performed based on VGG-16, and we compare corruption and perturbation robustness for MaxPool (used in original human-designed networks), MaxBlurPool \cite{zhang2019making} (an anti-aliasing MaxPool), and shape adaptors. All models are trained on the clean dataset.

\begin{figure}[ht!]
  \begin{adjustbox}{valign=t, minipage={.63\textwidth}}
    \input{image/perturbation.tex}
  \end{adjustbox}%
  \quad
  \begin{adjustbox}{valign=t,minipage={.35\textwidth}}
    \vspace{0.6cm}
    \centering
    \scriptsize
    \setlength{\tabcolsep}{0.15em}
    \renewcommand\theadset{\renewcommand\arraystretch{0.6}}
    \begin{tabular}{lccc}
      \toprule
    &  \makecell{MaxPool\\(Vanilla)}  & \makecell{Max\\BlurPool} & \makecell{Shape\\ Adaptors}\\
  \midrule
    \thead[l]{\scriptsize CIFAR-10 \\ {\tiny (Clean Err.)}} &  5.89    &   5.86    &   {\bf 4.65}     \\
    \thead[l]{\scriptsize CIFAR-10-C \\ {\tiny (mean CE)}} &   26.78     &  25.61     &  {\bf 23.02}     \\
    \midrule
    \thead[l]{\scriptsize CIFAR-100 \\{\tiny (Clean Err.)}} &   24.61    &  25.04      &  {\bf 20.84}     \\
    \thead[l]{\scriptsize CIFAR-100-C \\  {\tiny (mean CE)}} &   51.45    &   51.43    &  {\bf 48.42}  \\
      \bottomrule
      \end{tabular} 
  \end{adjustbox}
  \vspace{-1cm}
  \caption{Left: mean flip rate of CIFAR-10-P dataset with difficulty level 3. Right: clean error on CIFAR-10 dataset and mean corruption rate on CIFAR-10-C dataset. All results are trained with VGG-16 on the clean dataset.}
  \label{fig:robustness}
\end{figure}
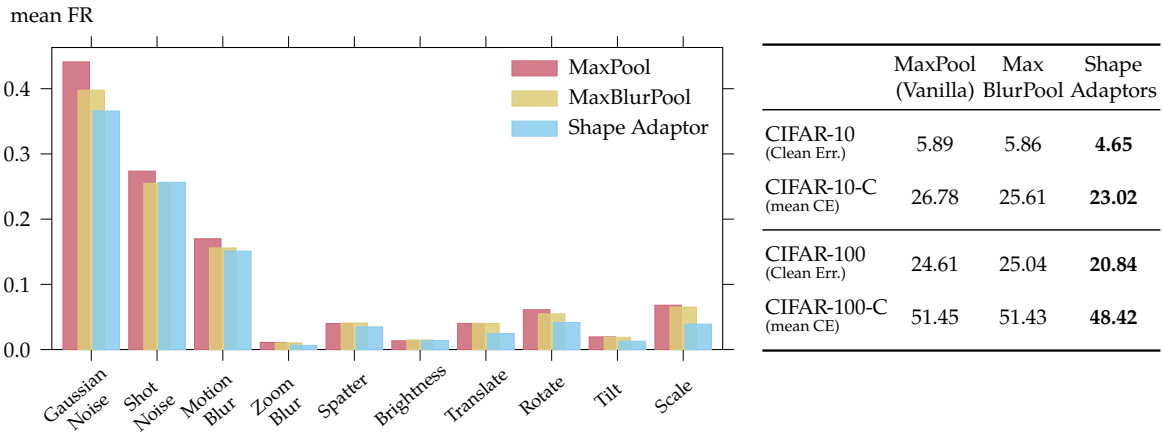

In Figure \ref{fig:robustness} Right, we can observe that shape adaptors equipped VGG-16 perform the best among all tested resizing modules by a large margin, in both corrupted and the clean datasets. In Figure \ref{fig:robustness} Left, we show that shape adaptors are able to improve almost every type of perturbation compared to the results from both MaxPool and the improved MaxBlurPool modules. This is most prominent in {\it digital type} perturbations (translate, rotate, tilt, scale), which provides approximately 40\% of performance improvements compared to MaxPool. These positive results show that shape adaptors not only can improve human-designed networks in accuracy, but also in robustness, by learning spatial-invariant features.

\section{Negative Results}
\begin{itemize}
  \item {\bf Other choices in shape adaptor search space.} We experimented with shape adaptors in the search space $\mathcal{R}=(0.25, 1)$, which we found to converge to a similar overall network shape, but with degraded performance compared to the current setting. We also experimented with shape adaptors using the search space in $\mathcal{R}=(0.5, 2.0)$, which we found to have very unstable learning dynamics, and often with out of memory issues.
  
  \item {\bf Other choices in reshaping function design.} We evaluated shape adaptors with its reshaping factor $s(\alpha)= \frac{1}{\alpha/r_1+(1-\alpha)/r_2}$, a weighted harmonic mean, which we found to have no improvements compared to the current setting.
  
  \item {\bf Other optimisation methods.} We experimented with updating network shape parameters and weight parameters based on a different sample in the training dataset, which we found to have a degraded performance compared to the current setting.
  
  \item {\bf Learning shape with prior structure knowledge.} We have experimented with directly replacing human-designed resizing layers with shape adaptors, which we found to have a minor effect on final performance compared to the current setting.
  
  \item {\bf Alternative shape adaptor design in a residual cell.} We have experimented with an alternate design of the residual cell, with a $[1\times 1]$ convolution layer as the identity branch, and with the weight layer as the resizing branch. The final performance with such a design achieved worse performance compared to the design defined in Fig. \ref{fig:structure}.
\end{itemize}

\end{document}

%% file: image/random_search.tex
% This file was created by tikzplotlib v0.9.0.
\begin{tikzpicture}

\begin{axis}[
every axis y label/.style={at={(current axis.north west)},above=2mm},
legend cell align={left},
legend style={fill=none, draw=none, at={(0,1)}, anchor=north west, font=\scriptsize},
log basis x={10},
tick align=outside,
tick pos=both,
x grid style={white!69.0196078431373!black},
xmin=250000000, xmax=11000000000,
xmode=log,
xtick style={color=black},
xtick={300000000,500000000,1000000000,2000000000,5000000000,10000000000},
xticklabels={300M,500M,1G,2G,5G,10G},
y grid style={white!69.0196078431373!black},
ymin=73.5, ymax=80.0,
ytick style={color=black},
ytick={74, 75, 76, 77, 78, 79, 80},
style={font=\small},
xlabel={MACs},
ylabel={Accuracy}
]
\addplot [only marks, draw=black, fill=white, mark size=2pt, colormap/viridis]
table{
x                     y
577578752 73.8627314567566
4491344384 79.1225096035004
};
\addplot [only marks, draw=black, fill=black, colormap/viridis, mark size=2pt]
table{
x                      y
604774592 74.0803062915802
497272256 74.1791903972626
439143936 74.2286324501038
466512896 74.2681980133057
391964672 74.4165360927582
574009280 74.426406621933
565462016 74.5450973510742
577595328 74.7428715229034
504156608 74.7626602649689
579189248 74.8022079467773
581453504 75.0988841056824
832702400 75.1977801322937
848832000 75.2373456954956
631960064 75.3362238407135
968106944 75.3461301326752
755905472 75.3658890724182
654265856 75.4351258277893
1428696576 75.4746973514557
857498048 75.5142331123352
907041984 75.5438983440399
1211623680 75.5735754966736
572906240 75.6329119205475
1258823680 75.7515728473663
718537664 75.781238079071
647978688 75.7911384105682
862568448 75.8010387420654
1453599232 75.8900463581085
1184801792 75.9097933769226
817622720 75.9394705295563
1448381184 76.0087072849274
1099469568 76.0185837745667
1559997952 76.0482490062714
1678775552 76.0482490062714
653268480 76.0581374168396
1245920256 76.0977149009705
1295745280 76.1768043041229
2435816448 76.1866986751556
691712192 76.2064814567566
718537664 76.2064814567566
1523280384 76.2361586093903
1799648256 76.2757062911987
960614336 76.2855887413025
1276356096 76.2955009937286
1188914432 76.3152778148651
950197952 76.3844907283783
1142982912 76.3844966888428
1154158080 76.4636039733887
2491441152 76.4734983444214
1307193856 76.4735043048859
2763203328 76.5427112579346
1946757888 76.5526235103607
2219855360 76.6218185424805
2942247936 76.6514956951141
1583696640 76.6515016555786
1515619328 76.6712844371796
1740182784 76.6910552978516
990610944 76.7108380794525
842932160 76.7405033111572
2772268544 76.7800688743591
1703992064 76.7899453639984
1399272960 76.7899572849274
1554917888 76.7998456954956
1357907200 76.8294990062714
652867008 76.8295049667358
1243481600 76.8393874168396
1276224256 76.8393933773041
1309330176 76.8393933773041
2179530240 76.8591642379761
2338374656 76.8690586090088
3355230464 76.9086182117462
1257335552 76.938271522522
1929239808 76.9976139068604
2496927488 77.0173907279968
1014580736 77.0174086093903
1896193536 77.0273149013519
1669204224 77.0866274833679
2858249984 77.0965099334717
1704170240 77.195405960083
3259643648 77.2053003311157
1998394368 77.2250711917877
1905556480 77.2448539733887
2044953088 77.2448778152466
1748813056 77.2547423839569
1062837440 77.2646307945251
1038409472 77.2844016551971
2190803456 77.2942900657654
4148934656 77.3734033107758
2560639744 77.3832976818085
3074800640 77.4030745029449
1879443456 77.4228572845459
2849143552 77.4228632450104
1863268352 77.4228692054749
1638935808 77.4327397346497
1178264064 77.4426341056824
2324558592 77.4722993373871
1627209984 77.5019705295563
1786874624 77.5118589401245
957734144 77.5118708610535
1701507328 77.5711894035339
2536147968 77.5711894035339
2349221888 77.5711953639984
1248029952 77.5909900665283
1311896832 77.660208940506
2037517056 77.6700794696808
2449235712 77.6700973510742
1527149312 77.6799857616425
2247843328 77.7492165565491
1767980544 77.7590930461884
2708673792 77.7689874172211
1201815552 77.7986407279968
1053680640 77.8382122516632
1946914048 77.8382182121277
3216778496 77.8481066226959
2623467264 77.9074549674988
2789156864 77.9173195362091
3946758144 77.9667556285858
1398066944 77.9667794704437
3778149888 77.976667881012
1446804992 77.9865562915802
3325812224 77.9964327812195
1338103040 78.0261099338531
3993064192 78.0261099338531
4740395520 78.0557751655579
1489541376 78.0656516551971
1774388224 78.0755519866943
8399946240 78.1150877475739
2863439360 78.1151115894318
5522093568 78.1348943710327
2172505856 78.144782781601
3160469248 78.1447887420654
8669726720 78.1645536422729
3580207104 78.2139778137207
3613078784 78.2140016555786
5657189888 78.2140016555786
4973794304 78.214031457901
3871545600 78.2634317874908
1310560000 78.3029854297638
5888508416 78.3030092716217
2320205056 78.3030152320862
3312406016 78.3128798007965
4800938496 78.3227741718292
5030080512 78.3227801322937
3136075008 78.332656621933
2236100352 78.3425509929657
2271929600 78.3524513244629
1981245440 78.3623337745667
4232977152 78.3722162246704
2423701248 78.3722281455994
4548935680 78.3722341060638
3041834240 78.3821225166321
3884092416 78.3821225166321
5127164928 78.4018933773041
3659904000 78.4315526485443
3111320832 78.4315645694733
7229920768 78.4414529800415
9650290688 78.4513235092163
2240134656 78.4513294696808
3907905536 78.4513294696808
4603295232 78.4611999988556
2249655552 78.461217880249
3546107136 78.461217880249
3791359232 78.4612238407135
1555579392 78.4612357616425
4156028160 78.4711122512817
2596663040 78.4810066223145
4899788288 78.4908950328827
1889166336 78.4909069538116
3255116032 78.5106599330902
1504822784 78.5205543041229
4398159872 78.5205662250519
2557526784 78.5304546356201
2403521024 78.5403370857239
2348549888 78.5502314567566
2505487104 78.5502374172211
4126429440 78.5601317882538
2961347328 78.5799026489258
9729015808 78.5897791385651
2432729600 78.589791059494
4205287936 78.6095559597015
2956375040 78.6490976810455
2807531520 78.6590039730072
4325513728 78.6590099334717
3138422528 78.6688983440399
3818577664 78.6689043045044
8506969600 78.6689043045044
3389653504 78.6787867546082
5728472064 78.6787867546082
4693189120 78.7084698677063
2564379648 78.7183463573456
2406147584 78.8370132446289
2957709056 78.8568019866943
2145043968 78.8765847682953
6680371712 78.8864612579346
2458574592 78.9655804634094
3777010176 78.9754748344421
4575002624 78.9952397346497
2672954368 79.0446937084198
2361085952 79.0941417217255
};
\addplot [only marks, draw=red, fill=red, mark size=2pt, colormap/viridis]
table{
x                     y
5210000000 79.16
};
\addplot [only marks, draw=orange, fill=orange, mark size=2pt, colormap/viridis]
table{
x                      y
314000000 75.39
};
\legend{Random Search (Best/Worst), Random Search, Shape Adaptors, Human Designed};
\end{axis}
\end{tikzpicture}

%% file: image/perturbation.tex
% This file was created by tikzplotlib v0.9.0.
\begin{tikzpicture}

\definecolor{color0}{RGB}{136, 204, 238}
\definecolor{color1}{RGB}{221, 204, 119}
\definecolor{color2}{RGB}{204, 102, 119}

\begin{axis}[
ybar,
every axis y label/.style={at={(current axis.north west)},above=2mm},
width=10.5cm,
height=5.6cm,
tick align=outside,
tick pos=both,
x grid style={white!69.0196078431373!black},
xmin=-0.6, xmax=9.6,
xtick style={color=black},
xtick={0,1,2,3,4,5,6,7,8,9},
xticklabel style = {rotate=40.0},
xticklabels={Gaussian Noise,Shot Noise,Motion Blur,Zoom Blur,Spatter,Brightness,Translate,Rotate,Tilt,Scale},
y grid style={white!69.0196078431373!black},
ymin=0, ymax=0.463155,
xticklabel style={align=center,text width=1.1cm, font=\notsotiny, yshift=2mm},
yticklabel style={font=\scriptsize},
y label style={align=left,},
ytick style={color=black},
ytick={0,0.1,0.2,0.3,0.4,0.5},
yticklabels={0.0, 0.1, 0.2, 0.3, 0.4, 0.5},
ylabel={\scriptsize mean FR},
legend entries={MaxPool,MaxBlurPool, Shape Adaptor},
legend style={align=left, draw=none, at={(1,1)}, anchor=north east, font=\scriptsize, column sep=0.2em},
legend cell align={left},
legend image code/.code={
    \draw [#1] (0cm,-0.1cm) rectangle (0.6cm,0.1cm);
},
]
\addplot +[bar shift=-.2cm, fill=color2, draw=color2, fill opacity=0.85] coordinates {
(0,0.4411)
(1,0.2736)
(2,0.1699)
(3,0.01093)
(4,0.03982)
(5,0.01346)
(6,0.0401)
(7,0.0611)
(8,0.01944)
(9,0.068)};
\addplot  +[bar shift=0, area legend, fill=color1, draw=color1, fill opacity=0.85]coordinates {
(0,0.3977)
(1,0.2548)
(2,0.1558)
(3,0.01022)
(4,0.04068)
(5,0.01472)
(6,0.0399)
(7,0.0547)
(8,0.01878)
(9,0.0651)};
\addplot  +[bar shift=+.2cm, area legend, fill=color0, draw=color0, fill opacity=0.85]coordinates {
(0,0.3658)
(1,0.2565)
(2,0.1509)
(3,0.00643)
(4,0.03475)
(5,0.01371)
(6,0.0245)
(7,0.0415)
(8,0.01268)
(9,0.03899)};
\end{axis}

\end{tikzpicture}

%% file: shape_adaptor_arxiv.bbl
\begin{thebibliography}{10}

\bibitem{brock2018smash}
Andrew Brock, Theo Lim, J.M. Ritchie, and Nick Weston.
\newblock {SMASH}: One-shot model architecture search through hypernetworks.
\newblock In {\em International Conference on Learning Representations}, 2018.

\bibitem{cai2018proxylessnas}
Han Cai, Ligeng Zhu, and Song Han.
\newblock Proxyless{NAS}: Direct neural architecture search on target task and
  hardware.
\newblock In {\em International Conference on Learning Representations}, 2019.

\bibitem{chen2017deeplab}
Liang-Chieh Chen, George Papandreou, Iasonas Kokkinos, Kevin Murphy, and Alan~L
  Yuille.
\newblock Deeplab: Semantic image segmentation with deep convolutional nets,
  atrous convolution, and fully connected crfs.
\newblock {\em IEEE transactions on pattern analysis and machine intelligence},
  40(4):834--848, 2017.

\bibitem{chollet2017xception}
Fran{\c{c}}ois Chollet.
\newblock Xception: Deep learning with depthwise separable convolutions.
\newblock In {\em Proceedings of the IEEE conference on computer vision and
  pattern recognition}, pages 1251--1258, 2017.

\bibitem{deng2009imagenet}
Jia Deng, Wei Dong, Richard Socher, Li-Jia Li, Kai Li, and Li~Fei-Fei.
\newblock Imagenet: A large-scale hierarchical image database.
\newblock In {\em 2009 IEEE conference on computer vision and pattern
  recognition}, pages 248--255. IEEE, 2009.

\bibitem{eitz2012humans}
Mathias Eitz, James Hays, and Marc Alexa.
\newblock How do humans sketch objects?
\newblock {\em ACM Transactions on graphics (TOG)}, 31(4):1--10, 2012.

\bibitem{goodfellow2013multi}
Ian~J Goodfellow, Yaroslav Bulatov, Julian Ibarz, Sacha Arnoud, and Vinay Shet.
\newblock Multi-digit number recognition from street view imagery using deep
  convolutional neural networks.
\newblock {\em arXiv preprint arXiv:1312.6082}, 2013.

\bibitem{guo2019spottune}
Yunhui Guo, Honghui Shi, Abhishek Kumar, Kristen Grauman, Tajana Rosing, and
  Rogerio Feris.
\newblock Spottune: transfer learning through adaptive fine-tuning.
\newblock In {\em Proceedings of the IEEE Conference on Computer Vision and
  Pattern Recognition}, pages 4805--4814, 2019.

\bibitem{han2015deep_compression}
Song Han, Huizi Mao, and William~J Dally.
\newblock Deep compression: Compressing deep neural networks with pruning,
  trained quantization and huffman coding.
\newblock In {\em International Conference on Learning Representations}, 2016.

\bibitem{he2015spatial}
Kaiming He, Xiangyu Zhang, Shaoqing Ren, and Jian Sun.
\newblock Spatial pyramid pooling in deep convolutional networks for visual
  recognition.
\newblock {\em IEEE transactions on pattern analysis and machine intelligence},
  37(9):1904--1916, 2015.

\bibitem{he2016deep}
Kaiming He, Xiangyu Zhang, Shaoqing Ren, and Jian Sun.
\newblock Deep residual learning for image recognition.
\newblock In {\em Proceedings of the IEEE conference on computer vision and
  pattern recognition}, pages 770--778, 2016.

\bibitem{hendrycks2018benchmarking}
Dan Hendrycks and Thomas Dietterich.
\newblock Benchmarking neural network robustness to common corruptions and
  perturbations.
\newblock In {\em International Conference on Learning Representations}, 2019.

\bibitem{hutter2019automated}
Frank Hutter, Lars Kotthoff, and Joaquin Vanschoren.
\newblock Automated machine learning-methods, systems, challenges.

\bibitem{jang2017categorical}
Eric Jang, Shixiang Gu, and Ben Poole.
\newblock Categorical reparameterization with gumbel-softmax.
\newblock In {\em International Conference on Learning Representations}, 2017.

\bibitem{KrauseStarkDengFei-Fei_3DRR2013}
Jonathan Krause, Michael Stark, Jia Deng, and Li~Fei-Fei.
\newblock 3d object representations for fine-grained categorization.
\newblock In {\em 4th International IEEE Workshop on 3D Representation and
  Recognition (3dRR-13)}, Sydney, Australia, 2013.

\bibitem{krizhevsky2009learning}
Alex Krizhevsky and Geoffrey Hinton.
\newblock Learning multiple layers of features from tiny images.
\newblock Technical report, Citeseer, 2009.

\bibitem{kuen2018stochastic}
Jason Kuen, Xiangfei Kong, Zhe Lin, Gang Wang, Jianxiong Yin, Simon See, and
  Yap-Peng Tan.
\newblock Stochastic downsampling for cost-adjustable inference and improved
  regularization in convolutional networks.
\newblock In {\em Proceedings of the IEEE Conference on Computer Vision and
  Pattern Recognition}, pages 7929--7938, 2018.

\bibitem{lee2016generalizing}
Chen-Yu Lee, Patrick~W Gallagher, and Zhuowen Tu.
\newblock Generalizing pooling functions in convolutional neural networks:
  Mixed, gated, and tree.
\newblock In {\em Artificial intelligence and statistics}, pages 464--472,
  2016.

\bibitem{li2019random}
Liam Li and Ameet Talwalkar.
\newblock Random search and reproducibility for neural architecture search.
\newblock {\em arXiv preprint arXiv:1902.07638}, 2019.

\bibitem{liu2018progressive}
Chenxi Liu, Barret Zoph, Maxim Neumann, Jonathon Shlens, Wei Hua, Li-Jia Li,
  Li~Fei-Fei, Alan Yuille, Jonathan Huang, and Kevin Murphy.
\newblock Progressive neural architecture search.
\newblock In {\em Proceedings of the European Conference on Computer Vision
  (ECCV)}, pages 19--34, 2018.

\bibitem{liu2018darts}
Hanxiao Liu, Karen Simonyan, and Yiming Yang.
\newblock Darts: Differentiable architecture search.
\newblock In {\em International Conference on Learning Representations}, 2019.

\bibitem{liu2017learning}
Zhuang Liu, Jianguo Li, Zhiqiang Shen, Gao Huang, Shoumeng Yan, and Changshui
  Zhang.
\newblock Learning efficient convolutional networks through network slimming.
\newblock In {\em Proceedings of the IEEE International Conference on Computer
  Vision}, pages 2736--2744, 2017.

\bibitem{louizos2018learning}
Christos Louizos, Max Welling, and Diederik~P. Kingma.
\newblock Learning sparse neural networks through $l_0$ regularization.
\newblock In {\em International Conference on Learning Representations}, 2018.

\bibitem{maddison2017concrete}
Chris~J Maddison, Andriy Mnih, and Yee~Whye Teh.
\newblock The concrete distribution: A continuous relaxation of discrete random
  variables.
\newblock In {\em International Conference on Learning Representations}, 2017.

\bibitem{maji13fine-grained}
S.~Maji, J.~Kannala, E.~Rahtu, M.~Blaschko, and A.~Vedaldi.
\newblock Fine-grained visual classification of aircraft.
\newblock Technical report, 2013.

\bibitem{mallya2018piggyback}
Arun Mallya, Dillon Davis, and Svetlana Lazebnik.
\newblock Piggyback: Adapting a single network to multiple tasks by learning to
  mask weights.
\newblock In {\em Proceedings of the European Conference on Computer Vision
  (ECCV)}, pages 67--82, 2018.

\bibitem{mallya2018packnet}
Arun Mallya and Svetlana Lazebnik.
\newblock Packnet: Adding multiple tasks to a single network by iterative
  pruning.
\newblock In {\em Proceedings of the IEEE Conference on Computer Vision and
  Pattern Recognition}, pages 7765--7773, 2018.

\bibitem{morgado2019nettailor}
Pedro Morgado and Nuno Vasconcelos.
\newblock Nettailor: Tuning the architecture, not just the weights.
\newblock In {\em Proceedings of the IEEE Conference on Computer Vision and
  Pattern Recognition}, pages 3044--3054, 2019.

\bibitem{nilsback2008automated}
Maria-Elena Nilsback and Andrew Zisserman.
\newblock Automated flower classification over a large number of classes.
\newblock In {\em 2008 Sixth Indian Conference on Computer Vision, Graphics \&
  Image Processing}, pages 722--729. IEEE, 2008.

\bibitem{pmlr-v80-pham18a}
Hieu Pham, Melody Guan, Barret Zoph, Quoc Le, and Jeff Dean.
\newblock Efficient neural architecture search via parameters sharing.
\newblock In Jennifer Dy and Andreas Krause, editors, {\em Proceedings of the
  35th International Conference on Machine Learning}, volume~80 of {\em
  Proceedings of Machine Learning Research}, pages 4095--4104,
  Stockholmsmässan, Stockholm Sweden, 10--15 Jul 2018. PMLR.

\bibitem{real2019regularized}
Esteban Real, Alok Aggarwal, Yanping Huang, and Quoc~V Le.
\newblock Regularized evolution for image classifier architecture search.
\newblock In {\em Proceedings of the AAAI conference on artificial
  intelligence}, volume~33, pages 4780--4789, 2019.

\bibitem{ronneberger2015u}
Olaf Ronneberger, Philipp Fischer, and Thomas Brox.
\newblock U-net: Convolutional networks for biomedical image segmentation.
\newblock In {\em International Conference on Medical image computing and
  computer-assisted intervention}, pages 234--241. Springer, 2015.

\bibitem{Saleh_Elgammal_2016}
Babak Saleh and Ahmed Elgammal.
\newblock Large-scale classification of fine-art paintings: Learning the right
  metric on the right feature.
\newblock {\em International Journal for Digital Art History}, Oct. 2016.

\bibitem{sandler2018mobilenetv2}
Mark Sandler, Andrew Howard, Menglong Zhu, Andrey Zhmoginov, and Liang-Chieh
  Chen.
\newblock Mobilenetv2: Inverted residuals and linear bottlenecks.
\newblock In {\em Proceedings of the IEEE conference on computer vision and
  pattern recognition}, pages 4510--4520, 2018.

\bibitem{Simonyan15}
Karen Simonyan and Andrew Zisserman.
\newblock Very deep convolutional networks for large-scale image recognition.
\newblock In {\em International Conference on Learning Representations}, 2015.

\bibitem{WahCUB_200_2011}
C.~Wah, S.~Branson, P.~Welinder, P.~Perona, and S.~Belongie.
\newblock {The Caltech-UCSD Birds-200-2011 Dataset}.
\newblock Technical Report CNS-TR-2011-001, California Institute of Technology,
  2011.

\bibitem{xie2017aggregated}
Saining Xie, Ross Girshick, Piotr Doll{\'a}r, Zhuowen Tu, and Kaiming He.
\newblock Aggregated residual transformations for deep neural networks.
\newblock In {\em Proceedings of the IEEE conference on computer vision and
  pattern recognition}, pages 1492--1500, 2017.

\bibitem{yu2014mixed}
Dingjun Yu, Hanli Wang, Peiqiu Chen, and Zhihua Wei.
\newblock Mixed pooling for convolutional neural networks.
\newblock In {\em International conference on rough sets and knowledge
  technology}, pages 364--375. Springer, 2014.

\bibitem{Yu2020Evaluating}
Kaicheng Yu, Christian Sciuto, Martin Jaggi, Claudiu Musat, and Mathieu
  Salzmann.
\newblock Evaluating the search phase of neural architecture search.
\newblock In {\em International Conference on Learning Representations}, 2020.

\bibitem{zeiler2013stochastic}
Matthew Zeiler and Robert Fergus.
\newblock Stochastic pooling for regularization of deep convolutional neural
  networks.
\newblock In {\em Proceedings of the International Conference on Learning
  Representation}, 2013.

\bibitem{zhang2019making}
Richard Zhang.
\newblock Making convolutional networks shift-invariant again.
\newblock In {\em International Conference on Machine Learning}, pages
  7324--7334, 2019.

\bibitem{zhu2020resizable}
Yichen Zhu, Xiangyu Zhang, Tong Yang, and Jian Sun.
\newblock Resizable neural networks, 2020.

\bibitem{zoph2017neural}
Barret Zoph and Quoc~V Le.
\newblock Neural architecture search with reinforcement learning.
\newblock In {\em International Conference on Learning Representations}, 2017.

\end{thebibliography}
